\documentclass{article}

\usepackage{arxiv}
\usepackage{amsmath}
\usepackage{float}
\usepackage{afterpage}
\floatstyle{plaintop}
\usepackage{amssymb}

\usepackage{multirow}

\usepackage[utf8]{inputenc} 
\usepackage[T1]{fontenc}    
\usepackage{hyperref}       
\usepackage{url}            
\usepackage{booktabs}       
\usepackage{amsfonts}       
\usepackage{nicefrac}       
\usepackage{microtype}      
\usepackage{cleveref}       
\usepackage{lipsum}         
\usepackage{graphicx}
\usepackage[numbers]{natbib}
\usepackage{doi}

\title{Uncertainty-aware transfer across tasks using hybrid model-based successor feature reinforcement learning}

\date{}

\author{ \hspace{1mm}Parvin Malekzadeh \thanks{Published in \textit{Neurocomputing} 530 (2023): 165-187} \\
	  University of Toronto\\
	\texttt{p.malekzadeh@mail.utoronto.ca} \\
	\And
	\hspace{1mm} Ming Hou \\
	Defence Research and Development Canada (DRDC) Toronto Research Centre
	\And
	\hspace{1mm} Konstantinos N. Plataniotis \\
	University of Toronto
}

\usepackage{float}
\usepackage{afterpage}

\restylefloat{table}
\usepackage{color}

\usepackage{bm}
\usepackage{amsmath,tabularx}
\usepackage{amsfonts}
\usepackage{subfigure}
\newcolumntype{P}[1]{>{\centering\arraybackslash}p{#1}}

\usepackage{algorithm, algpseudocode}
\newcommand{\myindent}[1]{
\newline\makebox[#1cm]{}
}

\usepackage{expl3}
\usepackage{environ}
\ExplSyntaxOn
\seq_new:N \g_appendices_seq
\NewEnviron{Appendix}{\seq_gput_right:No \g_appendices_seq \BODY}
\newcommand\AddAppendices{
  \appendix
  \seq_map_inline:Nn \g_appendices_seq {##1}
}
\ExplSyntaxOff

\usepackage{float}
\usepackage{afterpage}

\restylefloat{table}
\usepackage{color}
\usepackage{bbm}

\usepackage{bm}
\usepackage{amsmath,tabularx}
\usepackage{amsfonts}
\usepackage{subfigure}
\newcolumntype{P}[1]{>{\centering\arraybackslash}p{#1}}
\usepackage{algorithm, algpseudocode}

\makeatletter
\newcommand{\multiline}[1]{%
  \begin{tabularx}{\dimexpr\linewidth-\ALG@thistlm}[t]{@{}X@{}}
    #1
  \end{tabularx}
}
\usepackage{expl3}
\usepackage{environ}
\ExplSyntaxOn

\ExplSyntaxOff

\usepackage{array}
    \newcolumntype{P}[1]{>{\centering\arraybackslash}p{#1}}

\usepackage[page]{appendix} 
\usepackage{stfloats}
\usepackage{chngcntr}
\usepackage{graphicx} 
\usepackage{caption}
\usepackage{booktabs} 

\def\mS{\mathcal{S}}
\def\mA{\mathcal{A}}
\def\SR{\text{UaMB-SF}}

\def\k{_{k}}
\def\t{_{t}}
\def\nk{_{k+1}}
\def\nt{_{t+1}}
\def\pt{_{t-1}}
\def\i{^{(i)}}
\def\bt{\bm{\theta}}
\def\h{\bm{h}}
\def\H{\bm{H}}
\def\K{\bm{K}}

\def\G{\bm{G}}
\def\A{\bm{A}}
\def\M{\bm{M}}
\def\I{\bm{I}}
\def\T{\bm{T}}
\def\P{\bm{P}}

\def\q{\bm{q}}
\def\F{\bm{F}}
\def\v{\bm{v}}

\def\g{\bm{g}}
\def\X{\bm{X}}
\def\Y{\bm{Y}}

\def\n{\bm{n}}

\def\um{\bm{\mu}}
\def\m{\bm{m}}
\def\Sig{\bm{\Sigma}}

\def\s{\bm{s}}
\def\pk{_{k-1}}

\def\aL{\mathcal{L}}

\def\x{\bm{x}}

\def\y{\bm{y}}

\def\v{\bm{v}}
\def\S{\bm{S}}
\def\B{\bm{B}}

\begin{document}
\maketitle

\begin{abstract}
Sample efficiency, which refers to the number of samples required for a learning agent to attain a specific level of performance, is central to developing practical reinforcement learning (RL) for complex and large-scale decision-making problems. The ability to transfer and generalize knowledge gained from previous experiences to downstream tasks can significantly improve sample efficiency. Recent research indicates that successor feature (SF) RL algorithms enable knowledge generalization between tasks with different rewards but identical transition dynamics. It has recently been hypothesized that combining model-based (MB) methods with SF algorithms can alleviate the limitation of fixed transition dynamics. Furthermore, uncertainty-aware exploration is widely recognized as another appealing approach for improving sample efficiency. An agent can efficiently explore to better understand an environment by tracking uncertainty about the value of each available action.
Putting together two ideas of hybrid model-based successor feature (MB-SF) and uncertainty leads to an approach to the problem of sample efficient uncertainty-aware knowledge transfer across tasks with different transition dynamics or/and reward functions. In this paper, the uncertainty of the value of each action is approximated by a Kalman filter (KF)-based multiple-model adaptive estimation. This KF-based framework treats the parameters of a model as random variables.
 To the best of our knowledge, this is the first attempt at formulating a hybrid MB-SF algorithm capable of generalizing knowledge across large or continuous state space tasks with various transition dynamics while requiring less computation at decision time than MB methods. We highlight why previous SF-based methods are constrained to knowledge generalization across same transition dynamics, present our novel approach on a firm theoretical foundation, and design a set of demonstration tasks to empirically validate  the effectiveness of our proposed approach. The number of samples required to learn the tasks was compared to recent SF and MB baselines. The results show that our algorithm generalizes its knowledge across different transition dynamics, learns downstream tasks with significantly fewer samples than starting from scratch, and outperforms existing approaches. We believe that our proposed framework can account for the computationally efficient behavioural flexibilities observed in the empirical literature and can also serve as a solid theoretical foundation for future experimental work.
\end{abstract}


\section{Introduction} \label{sec:Intro}
\noindent Reinforcement learning (RL) provides powerful algorithms for sequential decision-making problems by designing an agent that interacts with an environment modelled as a Markov decision process (MDP). The interaction between an agent and its environment consists of the agent selecting actions and the environment responding to those actions by presenting new situations (states) to the agent and generating rewards represented by special numerical values. The agent chooses its action based on a decision rule known as policy and attempts to learn the optimal policy that maximizes the value function, which is the sum of rewards over time.
%
The number of samples (i.e., interactions) with the environment required for an RL agent to learn a task can be enormous. {In such cases, the sample inefficiency of RL training leads to substantial computational costs, which prevents industrial applications from adopting it~\cite{buckman2018sample,zhou2019efficient}. For instance, AlphaStar~\cite{vinyals2019grandmaster} is an RL agent that can outperform the average ranked human player in all StarCraft II games. Nonetheless, the number of samples required is typically several orders of magnitude greater than that of a typical human player. Thus, replicating the results of AlphaStar would cost millions of dollars.~\cite{agarwal2022provable}.}

The scale of methods that can improve the sample efficiency of an RL algorithm is broad. However, in this work, we focus on two critical aspects that a sample efficient RL agent must address~\cite{osband2018randomized}: \textit{transfer learning} and \textit{exploration}. Transfer learning~\cite{peirelinck2022transfer, atkinson2021pseudo, transfer_survey, kirk2021survey} is the generalization and adaptation of an agent's prior knowledge from a seen (source) task to learn a downstream (also referred to as target or test) task resulting from changes in the seen task's environment without learning from exhaustive interactions from scratch~\cite{taylor,zanette2020learning, ding2019improved}. {The main idea of transfer learning is to construct a learning agent that does not start over every time it faces a new target task, but the agent can reuse its experience from the source task to improve its sample efficiency on the target task. For example, consider a robot that has been given the task of retrieving an object from a collapsed building, a situation in which it cannot receive direct supervision from a human. During the test-time trial, the robot must retrieve this object with the fewest samples while avoiding unknown obstacles. It may be able to complete this task using its knowledge of the building before the disaster.\\
Exploration is defined as the efficient exploration of unknown environments and the collection of informative experiences that can guide policy learning most effectively towards optimal ones~\cite{yang2021exploration,agarwal2021deep}.} In other words, an agent should not be stuck in what it knows about the environment because that information may be suboptimal. To find truly optimal behaviour, the agent must investigate actions about which it is unsure~\cite{menard2021fast}.

Transfer learning in the context of RL can be classified based on the RL frameworks used to learn a specific task. Traditionally, RL algorithms are categorized into \textit{model-free (MF)}~\cite{Parvin:access,Hu, count, huang2021going} and \textit{model-based (MB)}~\cite{2,3}. MF approaches learn cached value functions through trial and error interactions with the environment without access to the internal model of the environment, i.e., the MDP's reward function and transition dynamics. MF methods enable quick and computationally efficient action selection at decision time. However, as recent research~\cite{sabatelli2021transferability, chen2021improving, tao2021repaint, wang2020target} has shown, the adaptability of MF frameworks to environmental changes does not appear to be promising. This is due to the fact that an MF agent cannot adapt cached values of all states to changes in the environment. On the other hand, MB methods learn the environment model and then use dynamic programming algorithms~\cite{bellman1966dynamic, wang2022intelligent, duan2022adaptive} to find the optimal policy.  The transfer of knowledge based on MB methods results in the adaptation of the agent's behaviour to changes in the reward function or/and transition dynamics~\cite{ayoub2020model,huang2021adarl,eysenbach2020off, sasso2021fractional}. However, this flexibility is slow and computationally expensive.
\\
Building upon the successor representation (SR)~\cite{Dayan}, which encodes future state visitation frequencies in a small state space environment, an alternative approach has been proposed for transfer learning. The SR encodes the dynamics of the environment; therefore, it can be constructed without knowledge of the reward function. SR methods compute the value function by taking a single dot product of the SR and the reward value. As a result, SR methods fall between MF and MB categories because they provide adaptive behaviour of MB agents in the face of changes in the reward function, as well as efficient computation of MF agents~\cite{Brantley, vertes, Sam}.
\citet{Barreto} extended the SR to large and continuous state spaces by representing each state of the environment with a feature vector. In this context, the SR is known as the successor feature (SF) because it represents expected feature values rather than individual state visitation frequencies. The SR and SF, like the value function in the MF context, obey fixed-point equations that can be calculated using the temporal difference (TD) method. \citet{Barreto} also demonstrated that the SR and SF, when learned through the TD scheme, can be reused across tasks with different reward functions but the same transition dynamics. In this paper, the frameworks that use TD learning to learn the SR and SF are referred to as temporal difference successor representation (TD-SR) and temporal difference successor feature (TD-SF) algorithms, respectively.\\
 Numerous extensions of TD-SF algorithms for large and continuous action and state space problems have recently been proposed~\cite{Sam, Chen, kulkarni, SU, reinke2021xi, Hansen2020FastTI,siriwardhana2019vusfa}. { However, if the transition dynamics differ between tasks, these methods cannot generalize their knowledge.
Failure to do so prevents current SF-based transfer learning approaches from being used in real-world problems where transition dynamics and environment rewards may vary across environments of different but related tasks.  To the best of our knowledge, there has been little research on this topic. \citet{zhang2017deep} attempted to address this issue by considering a linear relationship between transition dynamics (and thus the SFs) of source and target tasks. However, this approach limits the application of transfer learning and is incapable of transferring and adapting knowledge between tasks with complex transition dynamics. Therefore, the problem of developing an SF-based method capable of generalizing its knowledge across tasks with varying transition dynamics remains unresolved.}

Several basic exploration techniques such as $\epsilon$-greedy and Boltzmann exploration~\cite{32, cesa2017boltzmann} have been developed in the RL literature. { These methods rely on point estimates of value functions and frequently use a random perturbation to initiate the exploration. Such algorithms, which rely on point estimates of value functions, cannot distinguish between an action that has never been chosen before (and thus requires exploration) and a suboptimal action that has been widely tried (and can be avoided). Moreover, in a sparse reward task, where an agent receives non-zero rewards at a meagre number of states, random exploration cannot learn the task in a reasonable amount of time and interactions. Thus, random exploration methods lead to undirected exploration, making the learning process slow and inefficient. As a result, other types of directed exploration are required. Count-based exploration~\cite{ostrovski2017count,machado2020count}, which directly uses state visitation counts, has been proposed to guide the agent toward underexplored states. Although count-based approaches are shown effective for exploration, they are prone to detachment and derailment~\cite{ecoffet2021first}. Detachment causes the agent to lose track of interesting areas to explore, and derailment  makes it difficult for the agent to return to previously visited areas. Count-based approaches are also notoriously short-sighted, causing the agent to become stuck in local minima~\cite{burda2018large}. A compelling alternative direction for exploration is to reduce uncertainty about the environment~\cite{agarwal2021deep,32, mavrin2019distributional, zhang2021towards, lockwood2022review, chua2018deep,zhou2020deep}. The uncertainty in this context refers to epistemic or parametric uncertainty~\cite{clements2019estimating}, which is provoked by the agent's imperfect knowledge of the environment given limited samples. When the agent is unsure about a specific area of the environment, it should not be overconfident and overuse its knowledge in that area. Instead, by interacting in that area, the agent should attempt to learn more about that area in order to reduce its uncertainty about it. While uncertainty-aware exploration has been considered in the context of several MF and MB works~\cite{chua2018deep, KOVA, maddox2019simple, azizzadenesheli2018efficient, likmeta2022directed, dong2021variance}, to the extent of our knowledge, there is very little research focusing on uncertainty-aware exploration within the transfer learning setting~\cite{balloch2022role}.}
%
\subsection{Contributions}
 { Due to complementary properties of MB and SF methods, i.e., flexible adaptation of MB algorithms following changes in either the reward function or transition dynamics and efficient computation of SF methods, \citet{russek2017, momennejad2017}, and~\citet{tomov2021multi} hypothesized that the brain uses both MB and SF methods in the form of parallel RL systems. A natural question arises from this hypothesis: is there a relationship between MB and SF methods, such that each field can learn from or potentially enhance the other? In this article, we present a novel model-based successor feature (MB-SF) approach in which an analytically principled and justified detailed connection between MB and SF algorithms is derived. Our proposed MB-SF algorithm uses the environment model, i.e., the reward function and transition dynamics, to compute the SF and value function. MB-SF removes the fixed transition dynamics assumption of current SF-based transfer learning methods and mitigates the high computational complexity of MB methods. The proposed MB-SF framework also benefits from desired features of MB and SF methods as it enables the agent to generalize its knowledge across tasks with various rewards and transition dynamics (similar to an MB agent) while requiring less computation than an MB agent (similar to an SF agent).
A common real-world RL use case for MB-SF will most likely be in scenarios where previously trained agents are available, making MB-SF an important research topic. Hence, MB-SF allows the broader community to tackle more complex RL problems without the need for a large number of samples or the computational resources required to train another agent from scratch on a new problem. Moreover, MB-SF produces behaviours that are considered signatures of decision-making in the empirical literature: immediate adjustment of knowledge in response to changes in the reward function or transition dynamics. Therefore, we believe MB-SF can provide a theoretical foundation for the observations in the empirical literature~\cite{momennejad2017}. }
%
%

As stated, MB-SF requires the environment model (i.e., the MB component of MB-SF) to construct the SF and value function. Many traditional algorithms have been designed for model learning in small and finite state space environments~\cite{moerland2020model}, where an agent can investigate received rewards on all states in the state space. However, experiencing all of the states and learning the exact model in many real-world problems with large or continuous state spaces is infeasible. { A natural solution to this issue is to approximate the model with parameterized functions. Neural networks are commonly used to learn model parameters~\cite{wang2022intelligent, Chen, hafner2019dream, hafner2019learning, wang2019approximate}. These approaches treat the reward function and transition dynamics as deterministic functions of some parameters and calculate point estimates of the parameters. Nevertheless, these methods cannot capture the agent's uncertainty about the estimated parameters, which is valuable information for exploration.
Moreover, failing to report uncertainty makes reproducing the results of a neural network difficult except under identical random conditions, potentially leading to a reproducibility crisis similar to that which plagues other fields~\cite{baker20161}. }
Several attempts (e.g., Bayesian neural networks and bootstrap) have been made to make neural networks uncertainty-aware~\cite{ vuong2019uncertainty, clavera2018model, zhou2020deep, gawlikowski2021survey, abdar2021review,agarwal2021deep}. However, these  methods necessitate an ensemble of neural networks, parameter sampling, and several runs for each sample. They are also prone to over-fitting on small datasets, leading to poor predictions far into the future~\cite{chua2018deep}. Uncertainty estimation with neural networks has, therefore, been exceptionally challenging. { Alternatively, authors in~\cite{deisenroth2011pilco,pan2014probabilistic,deisenroth2013gaussian,cutler2015efficient} proposed to treat model parameters as random variables.  They used Gaussian processes to approximate the environment model. These methods, however, require full planning until the time horizon, which results in high computational costs for large state and action spaces. Furthermore, Gaussian processes cannot handle the potential non-stationarity of the model due to policy or MDP changes.}
\\
\citet{Geist} used {unscented Kalman filter (UKF)}~\cite{wan2000unscented}, a nonlinear version of {Kalman filter (KF)}~\cite{KF}, for approximation of the value function and its uncertainty in the context of MF frameworks. KFs~\cite{KF} are recursive Bayesian approaches for estimating states using noisy observations in dynamic environments. The proposed UKF-based estimation by~\citet{Geist} alleviates the complete planning requirement of Gaussian processes and can also handle the possible non-stationarity of the model. Recently, \citet{KOVA} employed another nonlinear version of KF, the extended Kalman filter (EKF), for approximating the value function in an MF setting. EKF linearizes a nonlinear model using the first-order Taylor series and then estimates the model parameters with the general form of linear KF. Nonetheless, linearization errors occur because the higher-order terms of the expansion are ignored during the EKF linearization process. UKF eliminates the linearization error introduced by EKF by directly capturing the value and uncertainty of the model parameters using an unscented transformation. Furthermore, unlike EKF, UKF does not require the Jacobian matrix to be computed. UKF, on the other hand, significantly increases computation because it involves Cholesky decomposition. Additionally, both UKF and EKF require an accurate nonlinear model of the system~\cite{gao2017interacting}, which leads to unreliable estimates in complex environments due to a lack of precise knowledge about the proper behaviour of the underlying environment.
\\
Multiple-model estimation approaches~\cite{liu2020selective, AK4, valipour2021constrained, malekzadeh2022akf} overcome the limitations of EKF and UKF by employing a weighted sum of multiple filters with different models as possible candidates for an unknown model of an environment.
Taking advantage of the benefits of multiple-model estimation frameworks, in this paper, we develop a multiple-model adaptive estimation method~\cite{AK4, assa2018similarity, hashlamon2020new} for approximating the environment model, which will be used to compute the SF in the proposed MB-SF scheme. Multiple-model adaptive estimation linearizes the unknown and potentially nonlinear reward function and transition dynamics as a bank of parallel linear models, each of which is estimated by a KF. A probabilistically weighted combination of each KF estimate provides the final estimate in multiple-model adaptive estimation.  Such a probabilistic fusion assigns the highest probability to the most accurate KF and lower probabilities to the other KFs. {Multiple-model adaptive estimation addresses the reproducibility crisis of neural network estimation and helps to reduce the risks associated with reproducing results under varying random conditions. Furthermore, the uncertainty information provided by multiple-model adaptive estimation leads to  the development of a novel uncertainty-aware exploration method that directs the agent to choose the most informative actions. As demonstrated by the experimental results, combining this uncertainty-aware exploration with the proposed MB-SF significantly improves sample efficiency. }

To empirically demonstrate the success of our resulting algorithm, which we call Uncertainty-aware Model-Based Successor Feature ($\SR$), in improving the sample efficiency of an RL agent through MB-SF and uncertainty-aware exploration, we perform $\SR$ on two sets of commonly used tasks in SF-based transfer learning studies~\cite{vertes,lehnert2020}: (1) a $2$-dimensional continuous navigation task, and (2) a revision of the $2$-dimensional discrete combination lock task designed in~\cite{lehnert2020}. Both the navigation and combination lock are challenging tasks for exploration since they have sparse reward functions; therefore, using no exploration scheme or random exploration is unlikely to lead to any rewards. {The experimental results are compared to the following recent approaches: AdaRL~\cite{huang2021adarl}, successor uncertainty (SU)~\cite{SU}, and MB Xi~\cite{reinke2021xi}. The results show that in contrast to SU and MB Xi, which only allow positive transfer learning across tasks with different reward functions, the proposed $\SR$ framework can generalize its knowledge across tasks with different rewards or/and transition dynamics while requiring fewer samples than AdaRL, which is an MB framework. }

In summary, by combining the ideas of a hybrid MB-SF method and an uncertainty-aware exploration via multiple-model adaptive estimation, we propose the $\SR$ framework, which makes the following contributions:
\begin{itemize}
\item {  Motivated by the efficient computation of SF methods and the flexible adaptability of MB algorithms in response to changes in reward function and transition dynamics, we propose a hybrid MB-SF approach that ties MB methods to SF algorithms and combines their complementary properties.   To the best of our knowledge, this is the first attempt to establish an analytical connection between MB and SF methods. The proposed MB-SF scheme removes limitations of previous SF-based transfer learning algorithms, generalizes its knowledge across variations in both reward function and transition dynamics, and contrasts with the computationally costly nature of MB methods at decision time.  MB-SF allows us to use the knowledge from the source task as a foundation for learning a new task with fewer data and computations than learning the task from scratch without any prior knowledge. As such, MB-SF is an attempt to lay the groundwork for the research workflow required for large-scale tasks in practical RL where previous computational work is available. Furthermore, MB-SF is a computationally efficient hypothetical mechanism for human and animal brains' flexible behaviour in response to environmental changes, and it can thus serve as a theoretical foundation for future empirical work~\cite{russek2017, momennejad2017, tomov2021multi}.}
\item{  Inspired by the recent success of uncertainty-oriented exploration in improving sample efficiency of RL algorithms~\cite{yang2021exploration,lockwood2022review}, we use an innovative method based on multiple-model adaptive estimation to approximate the environment model and the agent's uncertainty about it. We then incorporate estimated uncertainty about the approximated model with the MB-SF framework to derive a novel form of uncertainty-aware exploration. The proposed estimation method naturally accounts for uncertainty and deals with possible non-stationarity and nonlinearity of the model. These properties contrast with current approximation methods, which use neural networks to learn only point estimates of model parameters. Furthermore, despite neural networks that reproduce except under the exact same random conditions, estimating uncertainty in the proposed multiple-model adaptive estimation method allows it to produce reliable results under different random environmental conditions, particularly in real-world environments. Hence, the proposed multiple-model adaptive estimation algorithm can be regarded as a reliable alternative to neural network approximation.}
\item {We demonstrate the effectiveness of the MB-SF and uncertainty-aware exploration components of $\SR$ using two sets of tasks: one with continuous state space and another with large state space. These tasks help us present transfer learning setups more effectively across various reward functions or transition dynamics and have environments with intense exploration challenges due to the scarcity of their rewards. The empirical results demonstrate the interest and advantage of $\SR$ in improving both sample efficiency and computation efficiency of RL algorithms.}
\end{itemize}

The rest of the paper is structured as follows: Section~\ref{sec:related} provides an overview of related work in transfer learning in the RL domain. We describe some basic theoretical knowledge of RL  in Section~\ref{sec:formulation}. The proposed $\SR$ framework is developed in Section~\ref{sec:Our_SR}. Section~\ref{sec:generalization} provides the theoretical validation of the contributions of $\SR$, and Section~\ref{sec:results} presents experimental results.  The complexity of the proposed method is discussed in Section~\ref{sec:complexity}. Section~\ref{sec:disc} discusses the limitations of $\SR$ and future research directions, and finally, Section~\ref{sec:con} concludes the paper.

\section{Related work}
\label{sec:related}
{This section presents current transfer learning research in the domain of RL.
 While transfer learning is widely used by the supervised learning community~\cite{vandaele2021deep, ho2021evaluation, dominguez2019transfer}, it is still an emerging topic for RL algorithms~\cite{transfer_survey, Barreto}. Transfer learning can be more complicated in the context of RL due to challenges arising from the nature of RL, such as transferring knowledge in the context of an MDP rather than a stationary data
domain as in supervised learning, task diversity, and limited data sources due to rewards scarcity~\cite{balloch2022role}. 
 RL transfer learning methods can be broadly categorized according to algorithms used for learning tasks. Depending on the learning method, the types of tasks between which knowledge transfer is possible vary~\cite{transfer_survey}.
\subsection{Transfer learning in the context of model-free methods}
{MF frameworks are generally categorized into three main groups: (1) policy-based, (2) value-based, and (3) actor-critic~\cite{sutton2012}. Value-based algorithms learn the optimal value function and then use it to derive the optimal policy, whereas policy-based algorithms learn the optimal policy directly. Actor-critic methods~\cite{wang2019approximate, liu2021demonstration} are hybrid approaches that use policy-based methods to improve a policy while also evaluating it by estimating its corresponding value function.}
Several studies, including~\cite{sabatelli2021transferability, tyo2020transferable}, investigated the adaptability of value-based algorithms to environmental changes. They trained a value function on a source task and then transferred the value function's parameters to a new task that differed from the source task in transition dynamics. The findings revealed that transferring value function parameters learned through a value-based method results in negative knowledge transfer. Some other works, such as~\cite{chen2021improving,tao2021repaint, agarwal2021contrastive, rusu2016progressive}, investigated transfer learning in the context of policy-based and actor-critic techniques. They demonstrated that transferring knowledge in the form of policies across tasks with dissimilar rewards can be beneficial. Nevertheless, these methods necessitate a high level of computation since knowledge ensemble from multiple source tasks is required. Our proposed $\SR$ algorithm, on the other hand, only requires knowledge from a single source task.
\subsection{Transfer learning in the context of model-based methods} \label{sec:related_MB}
{ The first step in developing MB RL frameworks is to learn reward function and transition dynamics models. The learned models are then fed into dynamic programming algorithms like value iteration and policy iteration~\cite{wang2022intelligent, puterman2014markov} to iteratively derive an optimal policy without the need for interaction with the environment.} MB algorithms are hence considered more sample efficient than MF approaches. In this sub-section, we first discuss different model learning techniques in the literature and then present existing MB approaches used in transfer learning settings.}

\textbf{Model learning:}
Because of the Markovian property of states in an MDP, the reward function and transition dynamics depend only on local data~\cite{hasselt2012reinforcement}. As a result, the problem of learning an MDP model can be transformed into a standard supervised learning problem. Although learning an MDP model is not trivial, it is generally simpler than directly learning the value function or optimal policy in MF methods~\cite{chua2018deep, hasselt2012reinforcement}. However, it has been shown that a small error in the learned model can lead to a significant error in the value function estimate, making MB methods less competitive in their asymptotic performances than MF algorithms~\cite{sasso2021fractional}. To address this issue of MB schemes, early studies such as~\cite{deisenroth2011pilco,snelson2005sparse} proposed using Gaussian processes to capture the uncertainty of the learned model. However, the application of these algorithms is limited to tasks with small state and action spaces as they require full planning until the time horizon~\cite{fiedler2021practical}. Recently, neural networks have been used as nonlinear function approximations to learn the environment model~\cite{chua2018deep,hafner2019dream, hafner2019learning, vuong2019uncertainty, clavera2018model}. \citet{ha2018world} and~\citet{hafner2019learning}, for example, introduced world models that allow MB agents to learn policies in environments with high-dimensional continuous state spaces. They trained a variational autoencoder~\cite{kingma2019introduction, odaibo2019tutorial} to learn a compressed representation of states, which was then fed into a recurrent neural network to learn the model.
However, neural networks cannot provide reliable uncertainty estimates for network output and are commonly prone to over-fitting, resulting in poor asymptotic performance~\cite{agarwal2021deep}. Some works~\cite{buckman2018sample, chua2018deep, zhou2020deep, vuong2019uncertainty,  hafner2020mastering, paster2021blast} employed a bootstrapped ensemble of deep neural networks to predict the uncertainty of the approximated model and demonstrated that a pure MB approach could outperform an MF agent. { These methods require training on every single data sample, necessitating extensive computation and removing the benefit of uncertainty estimation via bootstrapping for action selection. Unlike the previous works mentioned, our multiple-model adaptive estimation method learns the environment model and its uncertainty while propagating the model uncertainty to the value function for more efficient learning.}

{ 
\textbf{Transfer learning based on the learned model:}
When applying transfer learning in the context of MB frameworks, in addition to the optimal policy or value function parameters learned in a source task, transition dynamics parameters can also be transferred to test tasks~\cite{eysenbach2020off}. The approaches proposed in~\cite{sasso2021fractional} and~\cite{landolfi2019model} learned a source task through a policy iteration algorithm and sequentially transferred the transition dynamics of the source task to several target tasks. When compared to learning the target tasks from scratch and random initialization, their experiments revealed a significant improvement in sample efficiency. Their methods, however, are limited to knowledge generalization across tasks with the same transition dynamics.
\citet{tang2021foresee} proposed decoupling the transition dynamics from the reward function so that they could be learned independently. This method necessitates the agent to learn an internal representation of the reward function, which must be relearned in order for the agent to adapt to a new reward function. \citet{agarwal2022provable, touati2021learning}, and~\citet{lu2021power}  proposed learning representations of transition dynamics of multiple tasks and then transferring the learned representations to a test task. Considering the test task representation as a mixture of the representations of the source tasks, an agent can immediately adapt to the test task by only learning the reward function. Such representation methods, however, can only generalize their knowledge to changes in the reward function.
\\
Recently, \citet{huang2021adarl} proposed AdaRL, an MB transfer learning approach for partially observable MDPs with MDPs as a special case. AdaRL learns the reward function and transition dynamics through a Bayesian neural network. The Bayesian neural network can be understood as introducing uncertainty into neural network weights, requiring parameter sampling and several feed-forward runs for each sample. Our method, however, avoids multiple parameter samples because uncertainty is propagated with every optimization step. AdaRL learns the value function through a value iteration algorithm. Their finding revealed that transferring  the reward function and transition dynamics parameters across tasks with various rewards or/and dynamics improves sample efficiency. Nonetheless, taking the learned model as a whole to compute the optimal value function through a value iteration method comes with high computational complexity. Our proposed framework instead adopts the learned model to construct the SF and then represents the value function as an efficient single dot product of the SF and the reward parameter. }
\subsection{ Transfer learning in the context of successor feature methods}
{Numerous extensions of SF RL have recently been proposed~\cite{tomov2021multi,lee2022toward}. Some directions include SF application to partially observable MDPs~\cite{vertes}, combination with max-entropy principles~\cite{vertes2020probabilistic}, or hierarchical RL~\cite{barreto2019option}. Methods that use SF algorithms for transfer learning are the most similar to our work.}
\citet{Chen} combined TD-SF with universal approximators to approximate the SF, which was then incorporated into actor-critic methods to facilitate learning new tasks with the same transition dynamics but different goals (i.e., different rewards) by transferring the SF and policy. 
\citet{kulkarni} combined a neural network with the TD-SF learning method to approximate the SF of a source task, which was then transferred to a new task with a different reward function. Other works on SF-based transfer learning include generalized policy improvement~\cite{barreto2018,barreto2020fast, ma2020universal, alegre2022optimistic,filos2021psiphi} and variational universal SFs~\cite{siriwardhana2019vusfa} that performed target driven navigation. These extensions are based on the assumption that the reward function is a linear combination of a fixed set of feature vectors. {\citet{reinke2021xi} proposed MB Xi, which extends the benefits of SF-based transfer learning to general nonlinear reward functions. This transfer learning method extends the SF to the $\epsilon$-function and learns it in a manner analogous to the TD-SF learning scheme.  Hence, MB Xi only applies to knowledge transfer across different rewards. To the best of our knowledge, our work is the first attempt to propose an SF-based transfer learning algorithm capable of generalization across tasks with various rewards or/and transition dynamics.}
\subsection{Uncertainty-aware transfer learning}
There are a few approaches in the context of transfer learning that allow dealing with both model approximation and model uncertainty at the same time.  { \citet{agarwal2022provable} proposed using an ensemble of learning networks to incorporate uncertainty into their MB transfer learning framework. SU~\cite{SU} and RaSF~\cite{gimelfarb2021risk} estimated uncertainty within the SF-based transfer learning domain using Bayesian linear regressions and optimizing entropic utilities, respectively. \citet{Geerts} and \citet{Mohammad:Icassp} incorporated KFs into TD-SF frameworks to derive the uncertainty of the SF within their SF-based transfer learning algorithms for finite state spaces and multiple agents problems, respectively. Recently, \citet{malekzadeh2022akf} combined multiple-model adaptive estimation with the TD-SF method to estimate the uncertainty of the approximated SF.  However, none of the preceding transfer learning frameworks can generalize their knowledge across different transition dynamics.}
%

\section{Background}\label{sec:formulation}
This section discusses the necessary background to follow the  developments presented in the rest of the paper. In what follows, we use the following notations: Scalar variables are represented by non-bold letters (e.g., $x$ or $X$), vectors by lowercase bold letters (e.g., $\x$), matrices by capital bold letters (e.g., $\X$), and the transpose of the matrix $\X$ is denoted by $\X^T$. Furthermore, all norms are assumed to be $L_2$.
\subsection{Reinforcement learning }
Consider an agent deployed in a stationary dynamic environment. The standard RL setting formalizes the environment as an MDP consisting of $5$-tuple $\{\mS, \mA, P, R, \gamma\}$, where $\mS$ and $\mA$ are the state space and action space, respectively. $P: \mS \times \mA \times \mS \rightarrow [0,1]$ shows the transition dynamics model, and $\gamma\in (0,1]$ is the discount factor, which controls the importance of future rewards versus the immediate ones. {$R: \mS \times \mA \rightarrow \mathbb{R}$ represents a bounded real-valued reward function, which is appropriately set up by an external supervisor. More specifically, the agent observes state $\s \in \mS$ and chooses action $a \in \mA$ based on a policy $\pi$, which can be deterministic or stochastic. A deterministic policy $\pi: \mS \rightarrow \mA $ certainly determines action $a$, whereas a stochastic policy $\pi: \mS \times \mA \rightarrow [0,1] $ specifies the probability of selecting action $a$. Immediately after performing action $a$ in state $\s$, the agent transits to state $\s' \in\mS$ with the probability of $P(\s'|  \s,a) $ and receives a reward $R(\s,a)$. } All transition probabilities for action $a \in \mA$ (i.e., $P(:|:, a)$) for MDPs with finite state and action spaces can be written in a stochastic matrix $\bm{P}^a$, such that the entry $(i, j)$ of the matrix $\bm{P}^a$ depicts the probability of transition to state $j$ immediately after taking action $a$ in state $i$. 
\\
Following a policy $\pi$, the value function $V^{\pi}(\s)$ and Q-value function $Q^{\pi}(\s,a)$ assign expected values to each state $\s$ and each state-action pair $(\s,a)$, respectively. The value function $V^{\pi}(\s)$ and Q-value function $Q^{\pi}(\s,a)$ under the policy $\pi$ are defined as
 \begin{eqnarray}
\lefteqn{\!\!\!\!\!\!\!\!\!\!\! V^{\pi}(\s)= \mathbb{E}_{P,\pi} \left[\sum_{k=t}^{k=\infty} \gamma^{k-t}  R(\s\k,a\k)|\s_t=\s \right],} \label{Eq:V}\\
&&\!\!\!\!\!\!\!\!\!\!\!\!\!\!\!\!\!\!\!\!\!\!\!\!  Q^{\pi}(\s,a)= \mathbb{E}_{P,\pi} \left[\sum_{k=t}^{k=\infty} \gamma^{k-t}  R(\s\k,a\k)| \s_t=\s, a_t=a\right],  \label{Eq:Q1} \\
&&\!\!\!\!\!\!\!\!\!\!\!\!\!  \s\k \sim P(.|\s\pk,a\pk), a\k \sim \pi(.|\s\k), \nonumber
\end{eqnarray}
where $t$ represents the current time step $t \in \{1,2,...  \}$, and the subscript of the expectation function $\mathbb{E}[.]$ denotes the probability distributions or density functions over which the expectation is computed. Since the reward function $R$ is assumed to be bounded, $V^{\pi}(\s)$ and $Q^{\pi}(\s,a)$ are well-defined functions. The objective in an MDP is to find an optimal policy $\pi^*$ that maximizes the value function $V^{\pi}(\s)$ or the Q-value function $Q^{\pi}(\s,a)$ for all states $\s \in \mS$.
\\
When the transition dynamics $P$ and the reward function $R$ are known, $Q^{\pi}$ can be recursively updated following the Bellman fixed-point equation as
{ \begin{eqnarray}
Q^{\pi}(\s,a)&\!\!\!=\!\!\!& R(\s,a) + \gamma \mathbb{E} _{ P} \left[V^{\pi}(\s\nt)\right], \label{Eq:Q} \\
&\!\!\!=\!\!\!& R(\s,a) + \gamma \mathbb{E} _{ P,\pi} \left[Q^{\pi}(\s\nt,a\nt)\right]. \label{Eq:Q-Q}
\end{eqnarray}}\normalsize
MB algorithms learn the environment model ( i.e., $P$ and $R$) and then compute the Q-value function through unrolling the recursion in Eq.~\eqref{Eq:Q-Q} into a series of nested sums, an algorithm known as value iteration.  MF algorithms, on the other hand, do not learn the environment model but instead use interaction samples to update estimates of the Q-value function or the policy. Given the transition data (sample) $\left< \s,a,\s', R(\s,a)\right>$ and  policy $\pi$, value-based MF algorithms update $Q^{\pi}(\s,a)$ through TD learning~\cite{32,KOVA, zhang2020deterministic} as:
\begin{eqnarray}
\!\! Q^{\pi}\nk(\s, a) = Q^{\pi}\k(\s, a) \, +  \alpha \Big(R(\s,a) +\gamma\,Q^{\pi}\k(\s', a')  - Q^{\pi}\k(\s, a) \Big),
\label{Eq:TD}
\end{eqnarray}
where $a' \sim \pi$ and ($0< \alpha \leq 1$) is the learning rate parameter. The term $R(\s,a) +\gamma\,Q^{\pi}\k(\s', a')  - Q^{\pi}\k(\s, a)$ is known as the TD error, which represents the difference between the predicted reward according to the current estimate of the Q-value function and the actually observed reward at iteration $k$.  The TD learning approach caches $Q^{\pi}$ estimates for all states $\s \in \mS$ and actions $a \in \mA$. Given the Q-value function for the policy $\pi$, we can find the optimal deterministic policy $\pi^*(\s)$ by choosing $a^*$ as $a^*=\pi^*(\s)= \arg\max_{b \in \mA} Q^{\pi}(\s,b)$.
%
\subsection{The successor representation  \label{sec:example}}
As an agent explores an environment, the states it visits are determined by the environment's transition dynamics and the policy. A representation that reveals the order of visiting states is likely to be more efficient for estimating the value function. \citet{Dayan} introduced the SR, an instance of such a representation that estimates the expected discounted sum of future occupancies for each state $\s''$ given the current state $\s$ and policy $\pi$:
{\begin{eqnarray}
{M}^{\pi}(\s,\s'') =\mathbb{E}_{P,\pi}
\left[\sum_{k=t}^{k=\infty}\gamma^{k-t}  \mathbbm{1}[\s\k=\s'']|\s_t=\s\right],\label{Eq:SR}
\end{eqnarray}}\normalsize
where $\mathbbm{1}\{\cdot\}=1$ if $\s\k=\s''$ and $0$ otherwise. To find ${M}^{\pi}(\s,\s'')$, an agent imagines starting a path from state $\s$ and then counts the number of times state $\s''$ will be visited subsequently. In a finite state space problem, the SR of all states can be shown in a matrix notation $\M^{\pi} \in \mathbb{R}^{|\mS| \times |\mS|} $, where $|\mS|$ is the cardinality of $\mS$.
Each entry ($i,j$) of $\M^{\pi}$ expresses the frequency of encountering state $j$ when starting a path at state
$i$ and following policy $\pi$.
Given the SR, the value function in Eq.~\eqref{Eq:V} can be expressed as the inner product of the SR and the immediate reward~\cite{Dayan}, i.e.,%
 \begin{eqnarray}
V^{\pi}(\s)=\sum_{\s''} {\M}^{\pi}(\s,\s'') r(\s'')
,\label{Eq:V_SR}
\end{eqnarray}
where $r(\s'')=\mathbb{E}_{\pi}[R(\s'',a'')]$. The most important benefit of SR-based frameworks is demonstrated by Eq.~\eqref{Eq:V_SR}, which represents an efficient linear mapping that allows the value function to be reconstructed straightforwardly based on the reevaluation of the reward function $R(\s'',a'')$ or the SR value $\M^{\pi}(\s,\s'')$ following changes in either the reward function or the SR. Given the transition data $\left<\s, a \rightarrow \s'\right>$ and the definition of the SR in Eq.~\eqref{Eq:SR}, a recursive formula of the SR is obtained as
\begin{eqnarray}
\bm{M}^{\pi}(\s,:)= \bm{1}_{\s}^T + \gamma \sum_{\s' \in S} \bm{T}^{\pi}(\s,\s') \bm{M}^{\pi}(\s',:)
,\label{Eq:SR_recursive}
\end{eqnarray}
 where $\bm{1}_s$ is a vector of all zeros except for a $1$ in the $\s^{\text{th}}$ position, and $\bm{T}^{\pi}$ is the one-step transition matrix for policy $\pi$, such that $\bm{T}^{\pi}(\s,\s') = \sum_{a \in \mA} \pi(a|\s) P(\s'|\s,a)$. As shown in Eq.~\eqref{Eq:SR_recursive},  the SR matrix $\bm{M}^{\pi}$ encodes aggregate state transition dynamics that are independent of the reward function $R$. Therefore, any changes in $P$ (and thus $T$) cause  the SR matrix $\bm{M}^{\pi}$ to change. 
 
 The main challenge here is determining how to learn the SR matrix so that its value will be immediately updated when the value of transition dynamics $P$ changes. The SR can be learned through the two main groups of TD-SR and MB-SR algorithms, which are discussed further below.
{\subsubsection{Successor representation learning using temporal difference method (TD-SR) }}
\citet{Gershman} indicated that, like the Q-value function in MF algorithms, the SR could be directly updated in a recursive form without the use of a one-step transition model $\bm{T}^{\pi}$, as follows:
\begin{eqnarray}
\!\!\!\!\!\!\!\!\!\! \M^{\pi}\nk(\s,:) = \M^{\pi}\k(\s,:) ~+   \alpha\Big(\bm{1}_{\s}^T +\gamma\,\M^{\pi}\k(\s',:) - \M^{\pi}\k(\s,:) \Big). \label{Eq:TD-SR}
\end{eqnarray}
In this paper, the methods that use Eq.~\eqref{Eq:TD-SR} to learn the SR are referred to as TD-SR algorithms. The TD-SR scheme caches a predictive representation of future states that aggregate over the one-step transitions; hence, it cannot adjust the entire SR matrix $\M^{\pi}$ in response to changes in transition dynamics at state $\s$. That is, TD-SR can adapt to changes in $P$ incrementally and through direct experiences. Therefore, TD-SR cannot immediately adjust the whole SR matrix to changes in the environment's transition dynamics  without experiencing new trajectories in full~\cite{momennejad2017}.
\vspace{.1in}{\subsubsection{Successor representation learning using transition dynamics model (MB-SR)} \label{sec:SR-MB11}}
Given the transition dynamics $P$ and policy $\pi$, the SR can be directly estimated via Eq.~\eqref{Eq:SR_recursive}. Since the transition dynamics model of the environment is used to learn the SR, this approach is known as MB-SR~\cite{russek2017}.  By unrolling the recursion in Eq.~\eqref{Eq:SR_recursive} into a series of nested sums, $\M^{\pi}(\s,:)$ can be rewritten as the sum of ($k-t$)-step transition probabilities starting from state $\s$ at time $t$:
{ \begin{eqnarray}
\label{Eq: Markov_SR}
 \M^{\pi}(\s,:) &\!\!\!=\!\!\!& \bm{1}^T_s + \gamma \T^{\pi}(\s,:)+ \gamma^2 (\T^{\pi})^2(\s,:)+ ...  \nonumber  \\
&\!\!\!=\!\!\!& \sum_{k=t}^{k=\infty} \gamma^{k-t}  \, (\T^{\pi})^{k-t}(\s,:)\textcolor{blue}{.}
\end{eqnarray}}\normalsize
The entire SR matrix $\M^{\pi}$ can be thus calculated as
{ \begin{eqnarray}
\M^{\pi} =\sum_{k=t}^{k=\infty} \gamma^{k-t} \, (\T^{\pi})^{k-t} = (\I - \gamma \T^{\pi})^{-1}{,} \label{Eq:SR_inverse}
\end{eqnarray}
}\normalsize
where $\I$ is an identity matrix. Since $\T^{\pi}$ is a stochastic matrix, $( \I - \gamma \T^{\pi})$ is invertible (proof is provided in the Appendix~\ref{sec:invert_A}).
\\
When the value of $P(\s'|\s,a)$ changes, the value of $\T^{\pi}(\s,\s')$ will be updated accordingly as $\bm{T}^{\pi}(\s,\s') = \sum_{a \in \mA} \pi(a|\s) P(\s'|\s,a)$. Eq.~\eqref{Eq:SR_inverse} then propagate the change in $\T^{\pi}(\s,\s')$ through the entire state space by taking the inverse of the matrix $(\I - \gamma \T^{\pi})$. This feature of MB-SR methods will be demonstrated further below with an illustrative example.
\vspace{.1in} \subsubsection{A motivating example:} \label{sec:moti_exam}
\begin{figure}[t]
\centering
\subfigure[ Up: Source task. Down: Test task created by placing a barrier in state $\s_4$ of the source task.]{
\label{subfig:source}
\includegraphics[scale=.43]{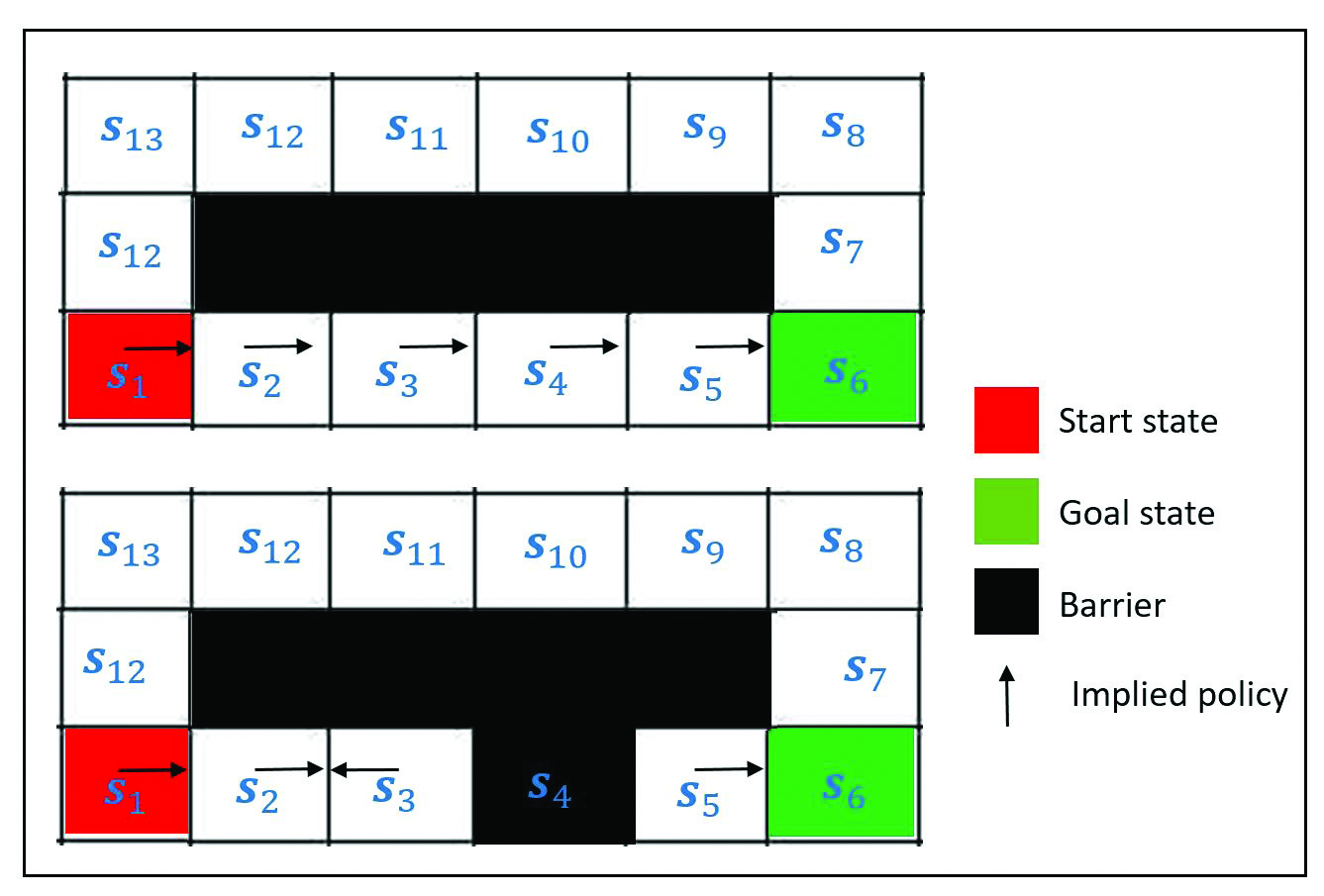}}
\\
\subfigure[ TD-SR can only update the SR value of a state following direct experiences with that state and its multi-step successors.]{
\label{subfig:test}
\includegraphics[scale=0.76]{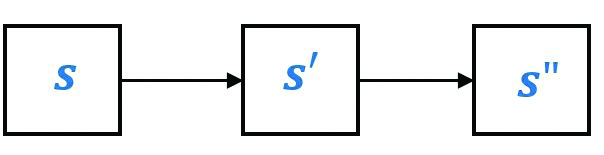}
}
\caption{An example for comparing the behaviours of TD-SR and MB-SR after changes in the transition dynamics $P$.}
\label{example}
\end{figure}

Let's consider the grid-world task shown in Fig.~\ref{subfig:source}. First, an agent learns the value function and optimal policy of the source task (Fig.~\ref{subfig:source}, up panel) by learning the reward function $R_{\text{source}}$ and the SR matrix through both the TD-SR ($\M^{\pi}_{\text{source: TD-SR}}$) and MB-SR ($\M^{\pi}_{\text{source: MB-SR}}$) methods. The learned $\M^{\pi}_{\text{source: TD-SR}}$, $\M^{\pi}_{\text{source: MB-SR}}$, and $R_{\text{source}}$ are then transferred to initialize $\M^{\pi}_{\text{test: TD-SR}}$, $\M^{\pi}_{\text{test: MB-SR}}$, and $R_{\text{test}}$ that will be trained on the test task (Fig.~\ref{subfig:source}, down panel). The test task is generated by placing a barrier at state $\s_4$ of the source task while retaining the reward values consistent with the source task, i.e., $R_{\text{source}}=R_{\text{test}}$. During the first trial of learning the test task, the agent is not aware of the barrier; hence, it mimics actions learned from the source task until it notices the block by repeatedly being dropped at $\s_3$ rather than $\s_4$ after taking the optimal action realized for $\s_3$ in the source task (i.e., moving right). Now, we investigate the behaviours of TD-SR and MB-SR methods in adapting the values of $\M^{\pi}_{\text{test: TD-SR}}$ and $\M^{\pi}_{\text{test: MB-SR}}$ after noticing that, contrary to the source task, $\s_4$ does not follow $\s_3$ in the test task:
\begin{itemize}
\item {\textbf{TD-SR}}: When the agent is dropped in  state $\s_3$, the row of the SR corresponding to state $\s_3$ (i.e., $\M^{\pi}_{\text{test: TD-SR}}(\s_3,:)$ will be updated by Eq.~\eqref{Eq:TD-SR}.
 However, the rows of $\M^{\pi}_{\text{test: TD-SR}}$ corresponding to other states, such as $\M^{\pi}_{\text{test: TD-SR}}(\s_1,:)$ and $\M^{\pi}_{\text{test: TD-SR}}(\s_2,:)$, remain unchanged (the same values as the source task). Hence, the agent cannot infer that states on the right side of the barrier no longer follow $\s_1$ and $\s_2$ through $\s_4$. This is because TD-SR can only update the SR value at the state directly facing the barrier, i.e., $\s_3$. 
 Therefore, the values of $V^{\pi}$ for other states do not change, causing the agent to select its learned actions  from the source task at these states (i.e., going right) and to always ends at $\s_3$ rather than attempting other actions on the first visit to $\s_1$ and $\s_2$.
\item {\textbf{MB-SR}}: When the agent notices the block, the value of $P(\s_4|\s_3,a)$ for $a=\text{`moving right'}$ is updated to zero; hence, $\bm{T}^{\pi}(\s_3,\s_4) =  \sum_{a \in \mA} \pi(a|\s_3) P(\s_4|\s_3,a) = 0$.  When $\M^{\pi}_{\text{test: MB-SR}}$ is recomputed by Eq.~\eqref{Eq:SR_inverse} with the updated $\T^{\pi}$, the whole matrix $\M^{\pi}_{\text{test: MB-SR}}$ will be updated accordingly. Thus, the rows of $\M^{\pi}_{\text{test: MB-SR}}$ corresponding to $\s_1$, $\s_2$ and $\s_3$ no longer predict future occupancies of the states on the barrier and its right side (e.g., $\M^{\pi}_{\text{test: MB-SR}}(\s_1,\s_5)=0$). Therefore, $V^{\pi}$ values computed based on the updated $\M^{\pi}_{\text{test: MB-SR}}$  immediately direct the agent to return to the start state $\s_1$ and select a new path.
\end{itemize}
%
{ \begin{table*}[tp]
\caption{Comparison of generalization properties and computation at the test task for different RL algorithms.}\label{Table:1}
\centering
\begin{tabular}{|P{1.3cm}|P{2.7cm}|P{2.7cm}|P{2.85cm}|P{2.9cm}|P{1.926cm}|}
 \hline
{ Model} & {Representation} & {Generalize to variation in rewards} &  {Generalize to variation in transitions} &  { Computation at test task} &  {Speed of generalization}
\\ \hline
  \multirow{2}{*}{MF} &  \multirow{2}{*}{ $Q$: Cached value} & \multirow{2}{*}{ no} &  \multirow{2}{*}{no} &  {Retrieve cached value, lowest cost} & -
\\ \hline
\multirow{2}{*}{ MB} & {$R$: Reward \newline $\T$: one-step transition matrix} & \multirow{2}{*}{yes} &  \multirow{2}{*}{yes} & {Iteratively compute values, highest cost} & \multirow{2}{*}{slow}
 \\ \hline
 \multirow{2}{*}{TD-SR} & {$R$: Reward \newline $\M$: cached SR} & \multirow{2}{*}{yes} & \multirow{2}{*}{ no} &  {Combine cached SR with TD, intermediate cost} & \multirow{2}{*}{fast}
\\ \hline
\multirow{2}{*}{\textbf{MB-SR}} & { $R$: Reward \newline $\M$: (computed from $\T$) } & \multirow{2}{*}{\textbf{ yes}} &  \multirow{2}{*}{\textbf{yes}} &  {Combine SR with MB, intermediate cost} & \multirow{2}{*}{\textbf{fast}}
\\ \hline
\end{tabular}
\end{table*} }
{As illustrated by the motivating example, unlike MB-SR, TD-SR cannot instantly adapt the SR values for states that do not directly experience changes in transition dynamics. Instead, TD-SR can only learn about changes in the transition structure incrementally; hence, it requires full trajectories through the state space to adapt the entire SR to the changes. Consider the start state $s$ and the two future states $s'$ and $s''$ depicted in Fig.~\ref{subfig:test}. If TD-SR discovers that $\s''$ no longer follows $\s'$, it will not be able to deduce that $\s''$ no longer follows $\s$, and thus cannot update the SR value corresponding to the start state $\s$.}

 Table~\ref{Table:1} summarizes the generalization properties and computation required at test tasks for various RL algorithms.
\subsection{The successor feature }\label{sec:SF}
Learning the SR for each state is impractical in environments with large or continuous state spaces. In such cases, each state $\s$ is mapped into a $L$-dimensional state feature vector $\bm{\phi}(\s) \in \mathbb{R}^{L}$ by a given function $\phi$.
In this setting, the SR is generalized to the SF~\cite{Barreto}, which is expected cumulative discounted of future state features as:
{ {\begin{eqnarray}
\m^{\pi}(\s)=\mathbb{E}_{P,\pi}
\left[\sum_{k=t}^{k=\infty}\gamma^{k-t} \bm{\phi}(\s\k)|\s_t=\s \right]. \label{Eq:SFs}
\end{eqnarray}}\normalsize
%

{\subsubsection{Successor feature learning using temporal difference method (TD-SF)}}
Like the SR, the SF can be learned using the TD learning scheme and the transition data $\left<\s, a \rightarrow \s'\right>$ as follows:
{\ \begin{eqnarray}
\!\!\!\!\!\!\!\!\!\!\!\!\!\!\! \m^{\pi}\nk(\s) &\!\!\! \leftarrow \!\!\!& \m^{\pi}\k(\s) \label{Eq:SF-TD22}
+\alpha\big(\bm{\phi}(\s)+\gamma \m^{\pi}\k(\s')-\m^{\pi}\k(\s)\big).
\end{eqnarray}}\normalsize
Algorithms that learn the SF through the TD scheme are called TD-SF frameworks.  TD-SF algorithms, like TD-SR methods, cannot be used to transfer knowledge between tasks with different transition dynamics because they cannot adapt the SF values of the entire state space to changes in the dynamics structure.
Because the SF and transition dynamics cannot be written in matrix forms in large and continuous state space environments, formulating an MB-SF approach to obtain a closed form solution similar to the finite state space as in Eq.~\eqref{Eq:SR_inverse} is non-trivial. In order to achieve this goal, we propose the $\SR$ framework in the following section.

In addition to the transition matrix $\T^{\pi}$, which is required  to learn the SR via MB-SR, the reward function $R$ is needed to compute the value function through Eq.~\eqref{Eq:V_SR}. Hence, these questions may arise: (1) what is the point of combining the SR (or SF) and MB methods? and (2) why not just use pure MB frameworks to compute the value function?     These questions are answered by noting that when using the SR, the value function  is efficiently computed as a single product between the SR and the reward function, whereas pure MB algorithms use Eq.~\eqref{Eq:Q}, which requires evaluating a set of typically intractable integrals, slowing generalization speed.
{\section{$\SR$: Uncertainty-aware Model-Based Successor Feature } \label{sec:Our_SR}}
%
{In this sub-section, we present our proposed transfer learning method, which we call $\SR$. $\SR$ is composed of three essential components: (1) \textit{a novel model learning algorithm}, which approximates the model of a large or continuous state space environment while providing uncertainty about the approximation, (2) \textit{a hybrid MB-SF framework}, which constructs the SF using the approximated model. MB-SF combines the flexibility of MB methods with the efficient computation of SF approaches, and 3) \textit{an uncertainty-aware exploration}, which leverages the approximated model's uncertainty to improve the sample efficiency of MB-SF.}
%

\vspace{.1in} \subsection{ Model learning\label{sec:MB}}
In this sub-section, we discuss the estimation procedures for the reward function and the transition dynamics of an MDP with a large or continuous state space. We assume that, given the $L$-dimensional feature vector $\{\bm{\phi}\t(\s)\}_{\s \in \mS}$ at time step $t$, the following functions govern the reward and transition dynamics:
{ \begin{eqnarray}
 R\t(\s,a) &  \!\!\!\!\! = \!\!\!\!\! & g (\bm{\phi}\t(\s), a),
\label{Eq:reward_function} \\
\mathbb{E}_{P}[\bm{\phi}\t(\s\nt)|\s\t=\s,a\t=a] &\!\!\!\!\! = \!\!\!\!\!& f(\bm{\phi}\t(\s), a). \label{Eq:transition_function}
\end{eqnarray}}\normalsize
To approximate the reward function and transition dynamics, we utilize function approximations $\tilde{g}$ and $\tilde{f}$ that fit the actual reward function $g$ and the transition function $f$ in the given transition data $\left< \bm{\phi}\t(\s),a,\bm{\phi}\t(\s'),R\t(\s,a)\right>$.
The majority of existing methods for such approximations with sequential data use neural networks~\cite{chua2018deep,zhou2020deep, hafner2019dream,  hafner2019learning}. However, these methods require storing all of the transition data along with the network parameters to later compute the error gradients in a backpropagation process.  As a result, implementing such techniques necessitates a significant amount of memory. Furthermore, neural network approximation methods, in general, cannot handle the model's possible non-stationarity and are prone to over-fitting on small datasets. Additionally, they cannot calculate the uncertainty of the approximated model, which is valuable information for exploration in RL problems~\cite{Geist}.
Filtering algorithms~\cite{AK1,AK2,AK4}, on the other hand, are efficient techniques that process sequential data based only on the last interaction. Such approaches eliminate the necessity of the learning process to record the entire history of data; therefore, they require less memory than neural networks. It has also been shown that applied filtering algorithms in RL domains~\cite{Parvin:access,KOVA, Geist,  malekzadeh2022akf} benefit us by handling non-stationarity and estimating the uncertainty of the approximation.
We thus use KF-based methods to approximate the transition function $f$ and the reward function $g$. The general KF formulation provides the best linear estimation in terms of mean square error~\cite{KF}. A brief outline of KF is provided in Section~\ref{sec:KF} of the appendix.  EKF and UKF~\cite{julier1997new} have been proposed as modified versions of KF for approximating parameters of a nonlinear model.  EKF and UKF estimation necessitate the calculation of the Jacobian matrix and the Cholesky decomposition, which incurs high computational costs.  Model mismatch errors in EKF and UKF can lead to the divergence of their solutions~\cite{yang2017comparison}.  EKF and UKF thus require an accurate system model, which is difficult to obtain in practice.  Multiple-model adaptive estimation~\cite{AK4, valipour2021constrained, assa2018similarity} has been proposed to alleviate the problems of EKF and UKF by approximating a nonlinear model as a fusion of a bank of linear KFs (each corresponding to a candidate for the system's unknown model). Multiple-model adaptive estimation computes the weight of each filter at a specific time using the probability that a hypothesized model is in effect. We develop an innovative method for learning the reward function $\tilde{g}$ and transition function $\tilde{f}$ within the proposed $\SR$ framework using multiple-model adaptive estimation.

{\subsubsection{ Reward learning}}
{As previously stated, the reward function can be approximated as a function $\tilde{g}$ of the feature vector $\bm{\phi}\t(\s)$ and action $a$ given the transition data $\left< \bm{\phi}\t(\s),a,\bm{\phi}\t(\s'),R\t(\s,a)\right>$.  Since complete information about the true model of the reward function is not available, we use multiple-model adaptive estimation to approximate the function $\tilde{g}$ as a fusion of $M_{\text{KF}}$ linear models $\{\tilde{g}\i\}_{i\in \{1,2,..., M_{KF} \}}$:
\begin{eqnarray}
\tilde{g} (\bm{\phi}\t(\s),a) \approx \sum_{i=1}^{M_{\text{KF}}} w\i\t \tilde{g}\i (\bm{\phi}\t(\s),a) = \sum_{i=1}^{M_{\text{KF}}} w\i\t {\bm{\phi}\t^T(\s)}(\bt^a\t)\i, \label{Eq:LAM2}
\end{eqnarray}
}%
 where $w\i\t \in \mathbb{R}$ and  $(\bt^a)\i \in \mathbb{R}^{L \times 1}$ are the probabilistic weight and the parameter vector of the $i^{\text{th}}$ model, respectively,  which must be estimated. To accomplish this, multiple-model adaptive estimation treats the received reward $R\t(\s,a)$ as a noisy measurement of the value of each function $\tilde{g}\i$:
\begin{eqnarray}
R\t(\s,a)
= \underbrace{{\bm{\phi}^T\t(\s)}}_{\h\t} (\bt^a\t)\i
+ N\t\i , \label{Eq:reward_update}
\end{eqnarray}
where $N\t\i \in \mathbb{R}$ represents the measurement (observation) noise corresponding to the $i^{\text{th}}$ KF and is represented as a zero-mean white Gaussian noise with the variance of $(P\t^{N})^{(i)} \in \mathbb{R}$. To estimate $w\i\t$ and $(\bt^a\t)\i$, an evolutionary equation for $(\bt^a\t)\i$ must be defined within each KF.
In general, the true evolution of $(\bt^a)\i$ over time cannot be obtained. However, following earlier works~\cite{Parvin:access, Geerts},  a heuristic evolution model according to Occam razor principle is adopted in this paper, such that passage of time increases uncertainty without changing the mean belief of the estimated $(\bt^a)\i$:
{ \begin{eqnarray}
(\bt^a\t)\i &\!\!\! =\!\!\! & \G\t ( \bt^a\pt )\i + \bm{\omega}\t\i \label{Eq:reward_predict},
\end{eqnarray}}\normalsize
where $\G\t$ stabilizes the filter, and $\bm{\omega}\t\i$ is the evolution noise, assumed to be a zero-mean white Gaussian noise vector with the covariance matrix $(\P^{\bm{\omega}}\t)\i \in \mathbb{R}^{L \times L}$. After the initialization step of the $i^{\text{th}}$ filter, the vector $(\bt\t^a)\i$ and its posterior covariance matrix $(\bm{\Pi}\t^a)\i$ are predicted as
{ \begin{eqnarray}
( \bt_{t|t-1}^a )\i &\!\!\!=\!\!\!& \G\t ({\bt}\pt^a )\i, \label{Eq:16}\\
(\bm{\Pi}^a_{t|t-1})\i &\!\!\!=\!\!\!& \G\t (\bm{\Pi}_{t-1}^a)\i \G\t^T+ (\P^{\bm{\omega}}\t)\i. \label{Eq:17}
\end{eqnarray}}\normalsize
The observed reward $R\t(\s,a)$ is then used to update estimates of each filter as follows:
{ {\begin{eqnarray}
!\!\! (\K\t^a)\i &\!\!\!\!=\!\!\!\!& (\bm{\Pi}_{t|t-1}^a)\i \h\t^T \big(\h\t (\bm{\Pi}^a_{t|t-1})\i \h\t^T +(P^{N}\t)\i \big)^{-1},\label{Eq:18}\\
\!\!\!\!\!\! \!\!\!\!\!\!\!\!\!\! ({\bt}^a\t)\i &\!\!\!\! = \!\!\!\!& (\bt^a_{t|t-1})\i + (\K\t^a)\i \big(R\t(\s,a) - \h\t ({\bt}_{t|t-1}^a)\i \big), \label{Eq:19}\\
\!\!\!\!\!\!\!\!\!\!\!\!\!\!\!\!  (\bm{\Pi}_{t}^a)\i &\!\!\!\! = \!\!\!\!& \big(\I - (\K\t^a)\i \h\t\big) (\bm{\Pi}_{t|t-1}^a)\i,\label{Eq:20}
\end{eqnarray}}\normalsize
where $(\K\t^a)\i$ is known as the gain of the $i^{\text{th}}$ KF.
 The outputs of all KFs are then averaged using their normalized weights:
\begin{eqnarray}
\!\!\!\!\!\!\!\!\!\!\!\!\!\!\!\!\!\!\!   \bt^a\t &\!\!\!\!\!\!=\!\!\!\!\!\!& \sum_{i=1}^{M_{\text{KF}}}w\i\t (\bt^a\t)\i, \\
\!\!\!\!\!\!\!\!\!\!\!\!\!\!\!\!\!\!\!  \bm{\Pi}^a\t &\!\!\!\!\!\!=\!\!\!\!\!\!& \sum_{i=1}^{M_{\text{KF}}} w\i\t \left( \left(\bm{\Pi}_{t}^a \right)\i  + \left( ({\bt}^a\t)\i-\bt^a\t \right) \left(({\bt}^a\t)\i-\bt^a\t \right)^T \right).
\end{eqnarray}
Weight $w\i\t$ associated with the $i^{\text{th}}$ filter at time $t$ is computed using the augmented likelihood function $\aL\i\t $ as
\begin{eqnarray}\label{eq:core}
w\i\t &\!\!=\!\!& \frac{w\i\pt\aL\i\t}{\sum_{j=1}^{M_{\text{KF}}} w^{(j)} \pt\aL^{(j)}\t},
\end{eqnarray}
where $\aL\i\t=\text{Pr} \left (R\t(\s,a)|({\bt}_{t|t-1}^a)\i,(P^N\t)\i \right)$.
Details on the calculation of the likelihood function are provided in Appendix~\ref{sec:MMAE}.
Fig.~\ref{Fig:MMAE} depicts the multiple-model adaptive estimation diagram  used for approximating the reward function $R(\s,a)$.
\\
For the sake of simplicity, we consider $M_{\text{KF}}=1$, i.e., a linear reward function. Linear function approximations have been widely studied in the RL domain, and their efficient approximations and sample efficiency have been demonstrated~\cite{zanette2020learning,jin2020provably,  yang2020reinforcement}. Moreover, the experimental results provided in Section~\ref{sec:results} demonstrate that $M_{\text{KF}}=1$ in $\SR$ provides good approximations. It is worth mentioning that the proposed $\SR$ framework can be easily extended to any nonlinear reward model by changing the value of $M_{\text{KF}}$. The multiple-model adaptive estimation structure for $M_{\text{KF}} \neq 1$ has been implemented in our previous studies on MF and TD-SF RL~\cite{Parvin:access,malekzadeh2022akf}, and the same benefits in terms of capitalizing on uncertainty estimation have been attained.
\begin{figure}[bp]
\centering
\includegraphics[scale=0.35]{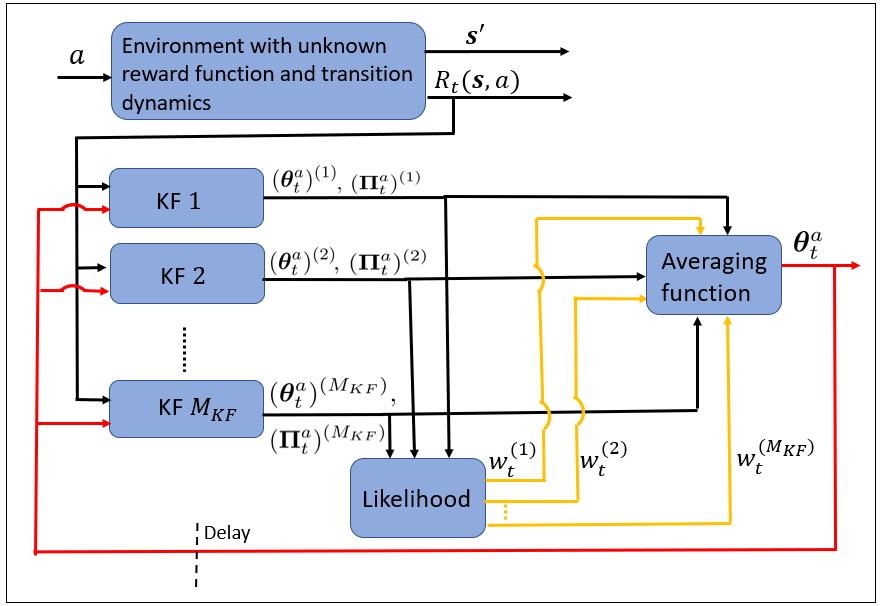}
\caption{ The structure of the multiple-model adaptive estimation with $M_{\text{KF}}$ parallel KFs used for reward function learning.}\label{Fig:MMAE}
\end{figure}

{\subsubsection{ Transition dynamics learning: }}
Similarly to the reward function approximation process, we approximate $f$ in Eq.~\eqref{Eq:transition_function} as a linear function of $\bm{\phi}\t(\s)$ (i.e., $M_{KF}=1$) with the parameter matrix $\F^a \in \mathbb{R}^{L \times L} $:
{ \begin{eqnarray}
\!\!\!\!\!\!\!\!\!\!\!\! \mathbb{E}_{P}[\bm{\phi}\t(\s\nt)|\s\t=\s,a\t=a]  \approx \tilde{f} = \F^a\t \bm{\phi}\t(\s). \label{Eq:LAM1}
\end{eqnarray}}\normalsize
The classical KF formulation (and thus multiple-model adaptive estimation) is developed to operate on scalar or vector parameters. One common method for estimating matrix variables such as $\F^a$ is to convert the matrix-based equations into a set of vector equations and then apply the conventional KF formulation, similar to the one used for estimating $\bt^a$.  This approach, however, causes estimation loss and generates extra equations for high-dimensional models due to the loss of the original structure. To tackle these problems, \citet{Matrix_kalman} proposed a matrix-based KF approach that directly provides optimal estimates in matrix format. We refer the reader to \cite{Matrix_kalman} for more details.  We use the matrix-based KF method with the following equations to estimate matrix $\F^a$:
 { \begin{eqnarray}
\F^a\t &\!\!\! =\!\!\! & 0.9 \F\pt^a + \bm{A}\t, \label{Eq:Matrix-KF}\\
\bm{\phi}\t(\s')&\!\!\!  =\!\!\! & \F^a\t \bm{\phi}\t(\s)+ \bm{B}\t,
\end{eqnarray}}\normalsize
where $\A\t$  is the evolution noise and $\bm{B}\t$ is the measurement noise, both of which are modelled as zero-mean white Gaussian  noises with the covariances of $\bm{\Sigma}^{\A}\t \in \mathbb{R}^{L^2 \times L^2}$ and $\bm{\Sigma}^{\B}\t \in \mathbb{R}^{L \times L}$, respectively. Like the classical KF, matrix-based KF estimates the value of matrix $\F^a\t$ along with its posterior covariance $\S\t^a$.

{\subsubsection{State feature learning:} }
So far, it has been assumed that the feature vector $\bm{\phi}(\s)$ is known a priori. This sub-section discusses an online learning process of $\bm{\phi}(\s)$ that satisfies Eqs.~\eqref{Eq:reward_update} and~\eqref{Eq:LAM1}. The feature vector $\bm{\phi}\t(\s)$ in this study is composed of $L$ radial basis functions (RBFs) as follows:
 \begin{eqnarray}
\bm{\phi}\t(\s) = \big[\phi^{(1)}\t(\s), \phi^{(2)}\t(\s),\ldots , \phi^{(L-1)}\t(\s), \phi^{(L)}\t(\s)\big]^T,\label{Eq:RBF}
\end{eqnarray} \normalsize
where $\phi^{(j)}\t(\s)$ is the $j^{\text{th}}$ RBF at time step $t$ and is modelled as a Gaussian with the mean $\um^{(j)}\t$ and the covariance $\bm{\Sigma}^{(j)}\t$:
 \begin{eqnarray}
\phi^{(j)}\t(\s) = e^{\frac{-1}{2}\left(\s-\um^{(j)}\t\right)^T\big({\Sig}^{(j)}\t\big)^{-1}\left(\s-\um^{(j)}\t\right)}. \label{Eq:phi}
\end{eqnarray} \normalsize
After initialization of $\{  \um^{(j)}_0 \}_{j=1:L}$ and $\{ \Sig^{(j)}_0 \}_{j=1:L}$, the loss function $J\t$ is defined to iteratively improve these parameters using stochastic gradient descent on the transition data $\left< \bm{\phi}\t(\s),a,\bm{\phi}\t(\s'),R\t(\s,a)\right>$:
{
\begin{eqnarray}
\!\!\!\!\!\!\!\! J\t &\!\!\!=\!\!\!& \left(R\t(\s,a) - (\bm{\theta}^a\t)^T \bm{\phi}\t(\s)\, \right)^2+ ||\bm{\phi}\t(\s')-\bm{F}^a\t \bm{\phi}\t(\s)||^2 + \left( ||\bm{\phi}\t(\s)||^2 -1 \right).  \label{Eq:loss}
\end{eqnarray}}\normalsize
The first and second terms of the equation above compute the approximation errors for the reward and transition dynamics, respectively.  The last term is a regularizer that necessitates the gradient to find the feature vector $\bm{\phi}(\s)$ with the unit norm. We empirically discovered that this term encourages stochastic gradient descent to learn an MDP model with $\arg\max_{a \in \mA} ||\bt\t^a|| \approx 1$ and $\arg\max_{a \in \mA} ||\F\t^a|| \approx 1$. It has been shown that in this setting, linear function approximations  provide near-optimal approximations of the reward and transition dynamics while reducing the number of samples~\cite{zanette2020learning, yang2020reinforcement, osband2014model}.
We assume that all constituent losses of $J\t$ contribute similarly to the overall loss $J\t$; therefore, the weighting parameter for each component is unnecessary. Moreover, to test the effect of the feature vector's dimension, i.e., $L$, on the approximated reward function and transition dynamics, this paper assumes that $L$ is a hyper-parameter.
{\subsection{Successor feature learning using the learned model (MB-SF)}} \label{sec:SFs learning}
This sub-section focuses on computing the SF using the previously learned transition matrix $\F^a\t$. Given an estimate of the feature vector $\bm{\phi}\t$, the SF can be computed using Eq.~\eqref{Eq:SFs} as
{ \begin{eqnarray}
\m^{\pi}\t(\s)&\!\!\!=\!\!\!& \mathbb{E}_{P,\pi}
\left[\sum_{k=t}^{\infty}\gamma^{k-t} \bm{\phi}\t(\s\k)|\s_t=\s \right] \nonumber
 \\
&\!\!\!=\!\!\!& \sum_{k=t}^{k=\infty} \gamma^{k-t} \mathbb{E}_{P,\pi} \left[ \bm{\phi}\t(\s\k) |\s_t=\s \right]  \nonumber \\
&\!\!\!=\!\!\!& \sum_{k'=0}^{k'=\infty} \gamma^{k'} \mathbb{E}_{P,\pi} \left[ \bm{\phi}\t(\s_{k'+t}) |\s_t=\s \right].  \label{Eq:SFs2}
\end{eqnarray}}\normalsize
According to the law of iterated expectations, $\mathbb{E}_{P,\pi} \left[ \bm{\phi}\t(\s\nt) |\s\t=\s \right]$ can be rewritten as
\begin{eqnarray}
\!\!\!\!\!\!\!\!\!\!\!\!\! \mathbb{E}_{P,\pi}\left[\bm{\phi}\t(\s\nt)|\s\t=\s \right] &=& \mathbb{E}_{\pi} \left[\mathbb{E}_{ P}[\bm{\phi}\t(\s\nt)|\s\t=\s,a\t=a] \right]  \nonumber  \\
\!\!\!\!\!\! &\approx & \mathbb{E}_{\pi}[\bm{F}^{a}\t \bm{\phi}\t(\s)]=\mathbb{E}_{\pi}[\bm{F}^{a}\t] \bm{\phi}\t(\s)= \bm{F}^{\pi}\t \bm{\phi}\t(\s),
\end{eqnarray}\normalsize
where $\mathbb{E}_{\pi}[\bm{F}^{a}\t]=\bm{F}^{\pi}\t=\sum_{a \in \mA} {\pi(a|\s) \F^a\t}$. For example, if policy $\pi$ selects actions uniformly at random, $\bm{F}^{\pi}$ is then calculated as $\bm{F}^{\pi}= \frac{1}{|\mA|}\sum_{a\in \mA} \bm{F}^a$, where $|\mA|$ is the cardinality of $\mA$.
\\
Starting from state $\s_t=\s$ and applying the dynamics and iterated expectations  law $k'$ times to the feature vector $\bm{\phi}\t(\s)$, $\mathbb{E}_{P,\pi} \left[ \bm{\phi}\t(\s_{k'+t}) |\s_t=\s \right]$ in Eq.~\eqref{Eq:SFs2} can be computed as
{
\begin{eqnarray}
\!\!\!\!\!\!\!\!\!\!\!\!\!  \!\!\!\!   \mathbb{E}_{P,\pi} \left[ \bm{\phi}\t(\s_{k'+t}) |\s_t=\s \right]\! &=& \!\mathbb{E}_{P} \bigg[ \mathbb{E}_{P,\pi} \left[ \bm{\phi}\t(\s_{k'+t}) |\s_{k'+t-1}, \s_t=\s \right] \!\! \bigg] \nonumber  \\
 \!\!\!\!\!\!\!\!\!\!\!\!  &=& \mathbb{E}_{P} \left[ \bm{F}^{\pi}\t \bm{\phi}\t(\s_{k'+t-1})|\s\t=\s  \right] = \bm{F}^{\pi}\t \mathbb{E}_{P} \left[   \mathbb{E}_{P,\pi} \left[  ... \right] \right] \nonumber \\
 \!\!\!\!\!\!\!\!\!\!\!\!  & =& (\bm{F}^{\pi}\t)^{k'}\bm{\phi}\t(\s).
\end{eqnarray}
} \normalsize
Therefore, the SF at time step $t$ is computed as:
{ \begin{eqnarray}
\!\!\!\!\!\!\!\!\!\!\! \m^{\pi}\t(\s)=\sum_{k'=0}^{k'=\infty} \gamma^{k'} (\bm{F}^{\pi}\t)^{k'} \bm{\phi}\t(\s) &\!\!\!\!=\!\!\!\!& \left(\I-\gamma\bm{F}^{\pi}\t \right)^{-1} \bm{\phi}\t(\s). \label{Eq:SR_LAM}
\end{eqnarray}} \normalsize
The SF formulation found in Eq.~\eqref{Eq:SR_LAM} is similar to the SR equation provided in Eq.~\eqref{Eq:SR_inverse} for finite state spaces. This formulation reduces to
the SR in the finite state space case, where $\bm{\phi}(\s)=\bm{1}_{\s}$.
%
\subsection{Uncertainty-aware exploration}
\label{sec:Active Learning}
{As previously stated,  the parameters $\bt\t^a$ and $\F\t^a$ were approximated using KF-based multiple-model adaptive estimation in order to benefit from the uncertainties of the approximations when developing an exploring policy. In this sub-section, we propose an uncertainty-based exploration method for instructing the MB-SF agent to select the most informative actions to speed up the learning process of a given task.
\\
Given the estimates $\bt\t^a$, $\m^{\pi}\t(\s)$, and $\bm{\phi}\t(\s)$ at time step $t$, the value function defined in Eq.~\eqref{Eq:V} can be calculated as follows
{
\begin{eqnarray}
 \!\!\!\!\!\! V^{\pi}\t(\s) &\!\!\!\!=\!\!\!\!& \mathbb{E}_{P,\pi} \big[\sum_{k=t}^{k=\infty} \gamma^{k-t} R\t(\s\k,a\k)| \s_t=s \big] \nonumber \\
 \!\!\!\!\!\!&\!\!\!\!=\!\!\!\!& \mathbb{E}_{P,\pi} \big[\sum_{k=t}^{k=\infty} \gamma^{k-t} \bm{\phi}^T\t(\s\k) \, \bt\t^a |\s_t=\s \big]   \nonumber  \\
 \!\!\!\!\!\! &\!\!\!\! \overset{(a)}{=} \!\!\!\!& \mathbb{E}_{P,\pi} \big[\sum_{k=t}^{k=\infty} \gamma^{k-t} (\bt^a\t)^T \bm{\phi}\t(\s\k)|\s_t=\s \big]  \nonumber  \\
 \!\!\!\!\!\! &\!\!\!\!\overset{(b)}{=} \!\!\!\!& \underbrace{ \mathbb{E}_{\pi}\left[(\bt^a\t)^T \right]}_{(\bt^{\pi}\t)^T } \underbrace{ \sum_{k=t}^{k=\infty} \gamma^{k-t} \mathbb{E}_{P,\pi} \left[ \bm{\phi}\t(\s\k) |\s_0=\s \right]}_{\bm{m}\t^{\pi}(\s)}  \nonumber  \\
 \!\!\!\!\!\! &\!\!\!\!=\!\!\!\!&  (\bt^{\pi}\t)^T \bm{m}\t^{\pi}(\s),
\end{eqnarray}}\normalsize
where (a) follows since $R\t(\s\k,a\k)$ is a scalar value, and (b) is because $\bt^a\t$ has no dependency on the transition dynamics $P$.
Using Eq.~\eqref{Eq:SR_LAM}, the value function can be computed as
{
\begin{eqnarray}
 V^{\pi}\t(\s) = \underbrace{(\bt^{\pi}\t)^T \left(\I-\gamma\bm{F}^{\pi}\t \right)^{-1}}_{(\v\t^{\pi})^T} \bm{\phi}\t(\s). \label{Eq:V_factor}
\end{eqnarray}}\normalsize
The equation above presents a computationally efficient mechanism (i.e., dot product) for scrutinizing the impact of changes in the reward function and transition dynamics on the value function through $(\bt^{\pi}\t)^T$ and $(\I-\gamma \bm{F}\t^{\pi})^{-1}$, respectively.
\\ Knowing the value function $V^{\pi}\t$, $Q^{\pi}\t(\s, a)$  can be obtained using the Bellman fixed point equation (i.e., Eq.~\eqref{Eq:Q}) as:
{
\begin{eqnarray}
\!\!\!\!\!\!\!\!\! Q^{\pi}\t(\s, a) \!=\! {(\bt^{a}\t)^T \bm{\phi}\t(\s) + \gamma \mathbb{E}_P [(\bt^{\pi}\t)^T \bm{m}\t^{\pi}(\s\nt)|\s,a]}. \label{Eq:Q-decision}
\end{eqnarray}}\normalsize
Based on Eqs.~\eqref{Eq:LAM1} and \eqref{Eq:SR_LAM}:
{
\begin{eqnarray}
\!\!\!\!\!\!\!\!\!\!\!\!\mathbb{E}_P [(\bt^{\pi}\t)^T \bm{m}^{\pi}\t(\s\nt)|\s,a] &\!\!\!\!\!=\!\!\!\!\!& (\bt^{\pi}\t)^T (\I-\gamma \bm{F}\t^{\pi})^{-1} \mathbb{E}_P[\bm{\phi}\t(\s\nt)|\s,a]\nonumber\\
\!\!\!\!\!\!\!\!\!\!\!\!\!\!\! &\!\!\!\!\!=\!\!\!\!\!& (\bt^{\pi}\t)^T (\I-\gamma \bm{F}\t^{\pi})^{-1} \bm{F}\t^{a} \bm{\phi}\t(\s).
\end{eqnarray}}\normalsize
By plugging the equation above into Eq.~\eqref{Eq:Q-decision}:
{ \begin{eqnarray}
Q^{\pi}\t(\s,a)= \underbrace{\left((\bt^a\t)^T + \gamma (\bt^{\pi}\t)^T (\I-\gamma \F\t^{\pi})^{-1} \F\t^a\right)}_{(\q^a\t)^T}  \bm{\phi}\t(\s). \label{Eq:factor}
\end{eqnarray}}\normalsize
Since $Q^{\pi}\t(\s,a)$ is a function of two action-dependent random variables $\bt^a\t$ and $\F\t^a$, its uncertainty can be calculated using uncertainties of these variables. We propose a learning policy that guides the agent to take the most informative action, leading to the most significant reduction in the uncertainty of the Q-value function.
Consider the current time step $t$, when the agent is in state $\s$ and must select an action $a \in \mA$. $\F^b\t$ and $\bt^b\t$ for $b \in \mA$ are not estimated since $R\t(\s,b)$ and $\s'$ have yet to be observed. However, their most recent estimates, namely, $\F^b\pt$ and $\bt^b\pt$, are available. Let $\sigma^2_ {Q\pt(\s,b)}$ to represent the variance of $Q^{\pi}\pt(\s,b)$. The agent thus chooses action $a$ according to the following deterministic policy:
{ \begin{eqnarray}
a =\pi\t(\s)= \arg\max_{b\in\mA} \left[Q^{\pi}\pt(\s,b) + \sigma_ {Q\pt(\s,b)}\right]. \label{Eq:action_var}
\end{eqnarray}} \normalsize
}
%
The variance of $Q^{\pi}\pt(\s,b)$ depends on the uncertainties of all random variables  $\bt\pt^b$, $\bt\pt^{\pi}$, $\F\pt^b$, and $\F\pt^{\pi}$. However, the choice of action $b$ only affects $\bt\pt^b$ and $\F\pt^b$. Uncertainties of the random variables $\bt^b\pt$ and $\bm{F}^b\pt$ are their posteriori covariances (i.e., $\bm{\Pi}^b\pt$ and $\S^b\pt$), which have been acquired by the KFs. Since traces of $\bm{\Pi}^b\pt$ and $\S^b\pt$ are the mean-square-errors of the approximations $\bt^a\pt$ and $\bm{F}^a\pt$, the most uncertain action is the one that has the largest approximation error. { Action $a$ is, therefore, selected as follows
{
\begin{eqnarray}
a = \arg\max_{b\in\mA} \left[Q\pt^{\pi}(\s,b) + \text{tr}\{\bm{\Pi}\pt^b+\S\pt^b\} \right], \label{Eq:action}
\end{eqnarray} }\normalsize
where tr\{.\} shows the trace operator of a matrix.}

{Fig.~\ref{fig:block} depicts the block diagram of the the proposed $\SR$ framework. Moreover, the $\SR$ algorithm is presented in Appendix~\ref{sec:Algorithm}.}
{
\begin{figure}[bp]
\centering
\includegraphics[scale=0.35]{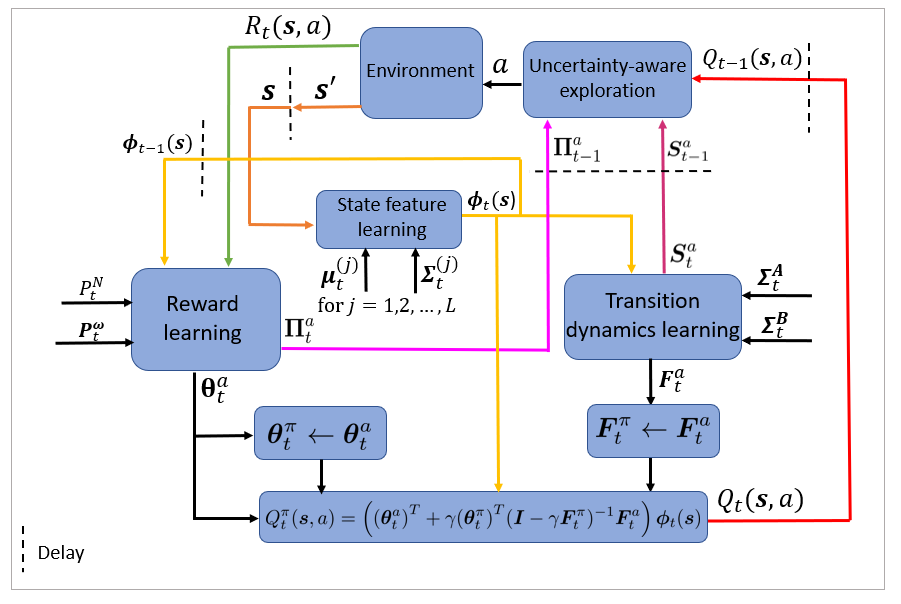}
\caption{ Visual Illustration of the proposed $\SR$ framework.}\label{fig:block}
\end{figure}
}
\subsubsection {Error bound for the approximated Q-value function} \label{sec:error_bound}
In view of the fact that we learned a linear approximation of the MDP model ($M_\text{\text{KF}}=1$), the obvious question that arises is: what if the approximation of the underlying model is not exactly accurate and thus misspecified? In this regard, we present an upper bound on the prediction error of the Q-value function approximated by Eq.~\eqref{Eq:factor} in terms of the one-step reward-prediction error $e^R\t=\sup_{\s \in \mS, a \in \mA} |R\t(\s,a)-(\bt^a\t)^T \bm{\phi}\t(\s)|$ and transition error $e^P\t=\sup_{\s \in \mS, a \in \mA} ||\mathbb{E}[\bm{\phi}\t(\s\nt)]-\F\t^a \bm{\phi}\t(\s)||$ (proof in Sub-section~\ref{sec:error_A} of the Appendix), as follows
{ \begin{eqnarray}
\!\!\!\! E\t=|Q\t^{\pi}(\s,a)-(\q^a\t)^T \bm{\phi}\t(\s)| \leq \frac{e\t^R+\gamma ||\v\t^{\pi}|| e^P\t}{1-\gamma}. \label{Eq:error_bound}
\end{eqnarray}}\normalsize
Under the linear approximation formulation and the assumption $||\bm{\phi}\t(\s)||=1$ in Eq.~\eqref{Eq:loss},  we have $\arg\max_{a \in \mA} ||\bt\t^a|| \approx 1$ and $\arg\max_{a \in \mA} ||\F\t^a|| \approx 1$. Thus, the approximation error bound obtained by Eq.~\eqref{Eq:error_bound} is close to optimal.
{\section{Theoretical validation of $\SR$ adaptability to environmental changes}
\label{sec:generalization}}
One essential contribution of the proposed $\SR$ algorithm is its ability to generalize and adapt its knowledge across tasks with different reward functions or/and transition dynamics.
 In a transfer learning setting, parameters learned in a source task are reused to initialize parameters in a new (test) task.
 In the transfer setting of our proposed $\SR$ framework, once $\SR$ learns a source task, the learned parameters $\{\bm{\phi}_{\text{source}}(\s)\}_{\s \in \mS}, \{\F^a_{\text{source}}, \S^a_{\text{source}}\}_{a \in \mA}$, $\{\bt^a_{\text{source}}, \bm{\Pi}^a_{\text{source}}\}_{a \in \mA}$, $\F^{\pi}_{\text{source}}$, and $\bt^{\pi}_{\text{source}}$ are transferred to initialize parameters $\{\bm{\phi}_{\text{test}}(\s)\}_{\s \in \mS}, \{\F^a_{\text{test}}, \S^a_{\text{test}}\}_{a \in \mA}$, $\{\bt^a_{\text{test}}, \bm{\Pi}^a_{\text{test}}\}_{a \in \mA}$, $\F^{\pi}_{\text{test}}$, and $\bt^{\pi}_{\text{test}}$ corresponding to $\SR$ implementation on a test task.
 \\
 We consider two possibilities: (i) the source task and test task differ in the reward value at the state-action pair $(\s',a') $, i.e., $R_{\text{source}} (\s',a')\neq R_{\text{test}}(\s',a')$, and (ii) the source task and test task vary in the transition dynamics $\left<\s',a' \rightarrow \s'' \right> $, i.e., $\F^{a'}_{\text{source}}\neq \F^{a'}_{\text{test}} $. The feature vectors of the source and test tasks are assumed to be the same in both scenarios, i.e., $\{\bm{\phi}_{\text{source}}(\s)= \bm{\phi}_{\text{test}}(\s) \}_{\s \in \mS}$.
Since $\SR$ selects actions using {Eq.~\eqref{Eq:action}}, the agent mimics actions learned in the source task on the first trial of learning the test task until it encounters the change at $\left<\s',a', R(\s',a'), \s'' \right>$.
\\ If the change is in the reward value (the former setting), the KF immediately adjusts the value of vector $\bt^{a'}_{\text{test}}$ and its covariance matrix ${\bm{\Pi}}^{a'}_{\text{test}}$, causing the update of $\bt^{\pi}_{\text{test}}$ as $\bt^{\pi}_{\text{test}} \leftarrow \bt^{a'}_{\text{test}}$. This change in the value of $\bt^{\pi}_{\text{test}}$, when combined with the matrix $(\I-\gamma \F^{\pi}_{\text{test}})^{-1}$ in Eq.~\eqref{Eq:factor}, causes the Q-value functions for all states $\s \in \mS$ in the test task to be updated. The agent then immediately adjusts its actions based on the new values of $Q^{\pi}_{\text{test}}$ and ${\bm{\Pi}}^{a'}_{\text{test}}$ using Eq.~\eqref{Eq:action}. \\
If the transition dynamics of the source and test tasks differ (the latter setting), the KF, which approximates transition dynamics, updates $\F^{a'}_{\text{test}}$ and its covariance matrix $\S^{a'}_{\text{test}}$. Updating $\F^{a'}_{\text{test}}$ causes $\F^{\pi}_{\text{test}}$ to be modified as $\F^{\pi}_{\text{test}} \leftarrow \F^{a'}_{\text{test}}$. The adjusted $\F^{\pi}_{\text{test}}$ will then be propagated throughout the entire state space by the term $(\I -\gamma \F^{\pi}_{\text{test}})^{-1}$ in Eq.~\eqref{Eq:factor}. Therefore, the value of $Q^{\pi}_{\text{test}}(\s)$ for each state $\s\in \mS$ instantly adapts to this change, and the agent modifies its actions by putting the new values in Eq.~\eqref{Eq:action}.

 The transfer setup of the proposed $\SR$ algorithm is presented in Fig.~\ref{fig:transfer}. $\bt^a$ and $\F^a$ capture changes in the reward function and transition dynamics, respectively. After training $\SR$ on a source task, only a few samples are needed to update $\bt^a$ and $\F^a$, eliminating the need for exhaustive interactions from scratch.
{
\begin{figure*}[bp]
\centering
\includegraphics[scale=.45]{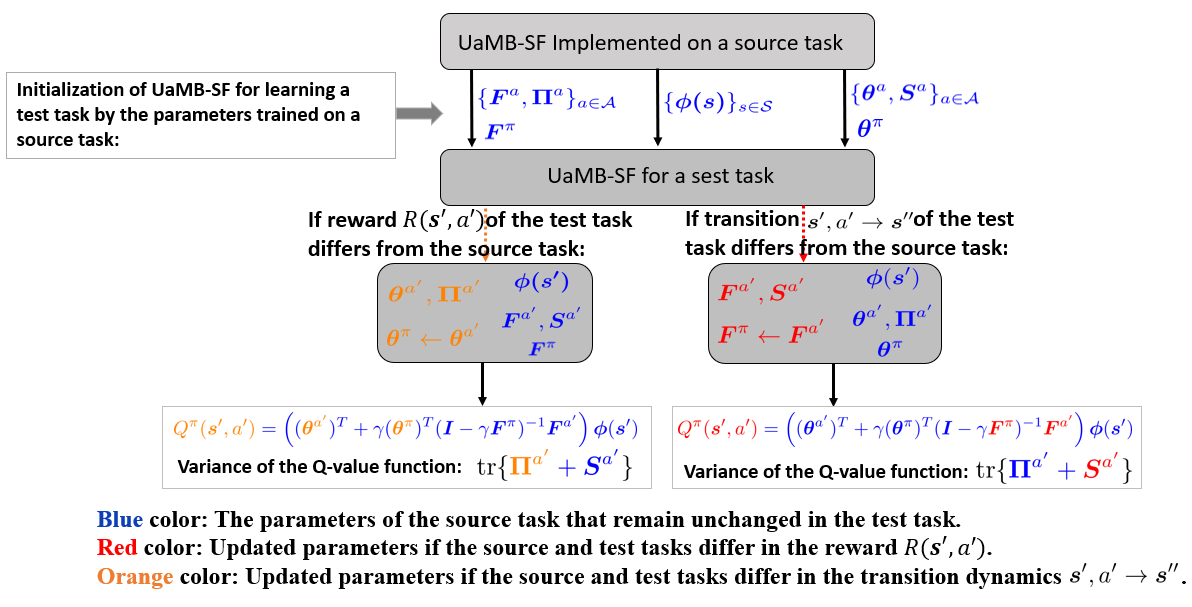}
\caption{ Transfer learning of the proposed $\SR$ framework: once the agent learns the source task, the trained parameters are kept and transferred to initialize the parameters of $\SR$ for learning the test task, which differs from the source task in  the reward or transition dynamics at $\left<\s',a', R(\s',a'),\s''\right>$. During the test task learning process, the agent mimics the actions learned in the source task until it encounters the change.}\label{fig:transfer}
\end{figure*}
}
\section{Experiments}
\label{sec:results}
{In this section, we conduct different sets of experiments to a) compare the performance of the proposed $\SR$ framework to existing algorithms when learning source tasks of reaching some goal states, b) show how transfer learning in $\SR$ facilitates learning of target tasks that vary from source tasks in reward values or/and transition dynamics, c) evaluate how the uncertainty-aware exploration and MB components of $\SR$ influence its performance, and d) indicate how variations in $\SR$'s hyper-parameters can affect its performance.}

{\textbf{Performance metric:} Different metrics can be used to evaluate the performance of RL algorithms, such as total accumulated rewards (total return) and episode length, which represents the number of training samples an agent used at each episode to achieve the goal. We are particularly interested in the episode length to measure an algorithm's adaptability to changes in a transfer learning setting because it has been widely used in state-of-the-art papers~\cite{agarwal2022provable, transfer_survey}, and more importantly, we cannot always expect return improvement, especially when the source and target tasks have very different reward functions~\cite{tao2021repaint}.}
\subsection{ Source tasks learning}
In this sub-section, we investigate the performance of $\SR$  for learning two goal-reaching source tasks and compare it to some recently published methods in the literature.

\textbf{Source tasks:}
We use the following tasks with sparse reward functions as source tasks in our experiments:
\begin{itemize}
\item[•] \textit{Continuous navigation task}, which  is a $2$-dimensional navigation task with states $\s=[x,y]^T \in [0,1]^2$, as shown in Fig.~\ref{subfig:taska}. Each episode starts from a uniformly sampled start state. The actions are to move left, right, up, or down and there is a $5\%$ chance that the agent will not move after selecting any actions.  The execution of each action moves the agent $0.05$ units in the desired direction. The agent cannot pass through the barrier. If the agent reaches one of the goal states, the reward is $+1$; otherwise, it is $0$.
\item[•]  \textit{Combination lock task}, which is a revision of the lock task in~\cite{lehnert2020}. Fig.~\ref{subfig:1} illustrates this task, in which an agent rotates the left and middle dials to obtain a rewarding number combination. The right dial is broken and whirls randomly at each time step. There are $216$ states $\s=[x_1,x_2,x_3]^T \in \{0,1,2,3,4,5\}^3$. Each episode begins with the left and middle dials set to $2$ and $4$, respectively. When the left and middle dials are set to $3$, a reward of $+1$ is given; otherwise, the agent receives no reward.
\end{itemize}
{
\begin{figure}[bp]
\centering
\subfigure[Source task A: $\s_b$ is the state where the agent must go around the barrier.]{
\label{subfig:taska}
\includegraphics[scale=0.34]{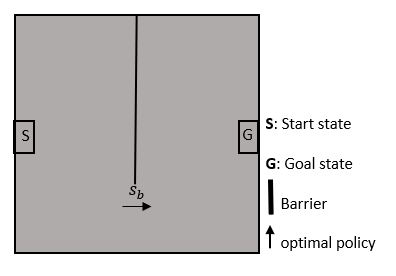}
}
\quad \quad \quad \quad \quad \quad
\subfigure[Source task 1: the right dial rotates randomly. ]{
\label{subfig:1}
\includegraphics[scale=0.37]{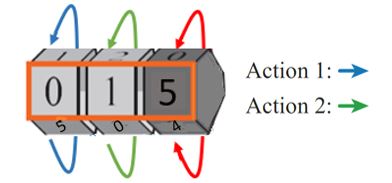}
}
\caption{Goal-oriented source tasks A and 1.}
\end{figure}
}
{
\textbf{Baselines:}
We compare the performance of the proposed $\SR$ framework to the following alternative algorithms on the continuous navigation and combination lock tasks:}
\begin{itemize}
\item[•] { \textit{SU~\cite{SU}:} SU establishes an uncertainty-aware transfer learning algorithm using the TD-SF scheme. It learns the SF through a neural network minimizing the TD error. To obtain the posterior distribution of the Q-value function, i.e., $\text{Pr}(Q)$, the weight vector $\bt^a$ is estimated using a Bayesian linear regression, which is a particular subset of KF. SU then selects actions that maximize a Q-value function sampled from $\text{Pr}(Q)$.  We use SU as a baseline to compare our hybrid MB-SF approach to TD-SF approaches.}
\item[•] {\textit{MB Xi~\cite{reinke2021xi}:} This is a transfer learning algorithm that eliminates the common assumption of linearly decomposable reward functions in most SF-based transfer learning algorithms. Instead of the SF, this work considers the $\epsilon$-function, which is the expected cumulative discounted probability of future feature vectors. MB Xi extends traditional SF-based transfer learning methods to a general reward function of feature vectors $\bm{\phi}(\s)$. It learns the reward function and transition dynamics through neural networks and then uses them to learn the $\epsilon$-function via a TD method analogous to TD-SF. Finally, the agent selects actions that maximize the Q-value function, which is computed as the dot product of the reward parameter and the $\epsilon$-function.  MB Xi is the most similar to our method in that it learns the environment model and then applies it to learn the $\epsilon$-function. The main differences between MB Xi and $\SR$ are as follows: a) MB Xi learns the $\epsilon$-function using the transition dynamics in the TD learning method, whereas $\SR$ learns the SF using the transition dynamics in Eq.~\eqref{Eq:SR_LAM}, and b) $\SR$ incorporates uncertainty for action selection, whereas MB Xi does not.} 
\item[•] {\textit{AdaRL~\cite{huang2021adarl}:} AdaRL is an MB transfer learning algorithm formulated in partial observable MDPs.  We consider the simplified AdaRL implementation for MDPs as we are interested in MDPs.  AdaRL learns the reward function and transition dynamics using Bayesian neural networks.  The agent then takes actions that maximize the Q-value function, which is approximated using a value-iteration algorithm. Therefore, by comparing AdaRL performance to that of SU, MB Xi, and $\SR$, we can compare the performance of pure MB transfer learning algorithms to that of SF-based and hybrid MB-SF transfer learning algorithms.}
\end{itemize}

{\textbf{Experimental setup:}
Let us represent the implementations of SU, MB Xi, $\SR$, and AdaRL on the source tasks as $\text{SU}_{\text{source}}$, $\text{MB Xi}_{\text{source}}$, $\SR_{\text{source}}$, and $\text{AdaRL}_{\text{source}}$, respectively. We initialize $\text{SU}_{\text{source}}$, $\text{MB Xi}_{\text{source}}$, and $\text{AdaRL}_{\text{source}}$ with the default hyper-parameters provided by their authors.  The values provided in Table~\ref{Table:param_nav} of the appendix are used to initialize the parameters of $\SR_{\text{source}}$. We train $\text{SU}_{\text{source}}$, $\text{MB Xi}_{\text{source}}$, $\SR_{\text{source}}$, and $\text{AdaRL}_{\text{source}}$ from scratch on task A across $500$ episodes, each with a maximum length of $200$ steps and on task 1 over $140$ episodes, each with a maximum length of $60$ steps. Each episode is terminated when the agent either achieves the goal or exceeds the maximum episode length.}

{\textbf{Results:}
Fig.~\ref{fig:experiment_source} compares the performance of SU, $\SR$, MB Xi, and AdaRL for learning the source tasks A and 1 from scratch. The solid lines and shaded area represent the average and standard deviation of episode length over $20$ independent runs. The shorter the episode, the sooner the agent can reach its goal. Table~\ref{Table:A} compares the average episode length across all episodes. $\SR_{\text{source}}$ and $\text{MB Xi}_{\text{source}}$  have roughly comparable performances because they both require learning the environment model, which is then used to approximate the SF and the $\epsilon$-function. $\text{SU}_{\text{source}}$ learns more quickly than other frameworks. This can be explained by the fact that SU only learns the reward parameter $\bt^a$, while $\SR$, $\text{MB Xi}$, and $\text{AdaRL}$ must approximate both the reward function and transition dynamics. Moreover, $\text{AdaRL}_{\text{source}}$ learns slower than $\SR_{\text{source}}$ and $\text{MB Xi}_{\text{source}}$. This follows because $\SR$ and $\text{MB Xi}$  compute the value function using a dot product, whereas $\text{AdaRL}$ approximates the value function iteratively through a value-iteration algorithm.}
{
\begin{figure}[bp]
\centering
\subfigure[Episode length for the source task A. ]{
\label{subfig:resulta}
\includegraphics[scale=.3]{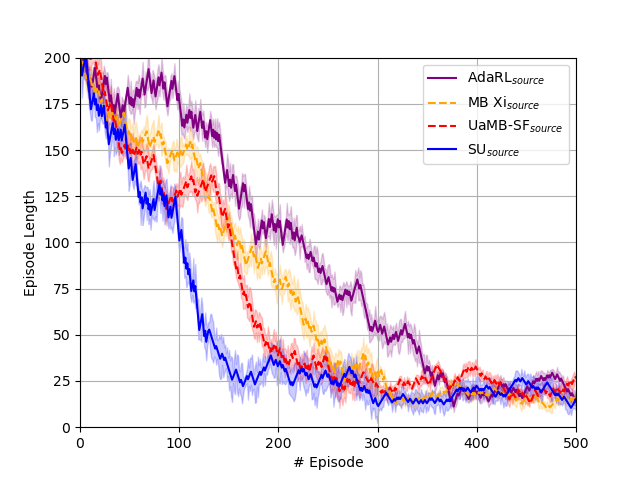}}
\quad \quad \quad \quad \quad \quad 
\subfigure[Episode length for the source task 1.]{
\label{subfig:steps1}
\includegraphics[ scale=0.3]{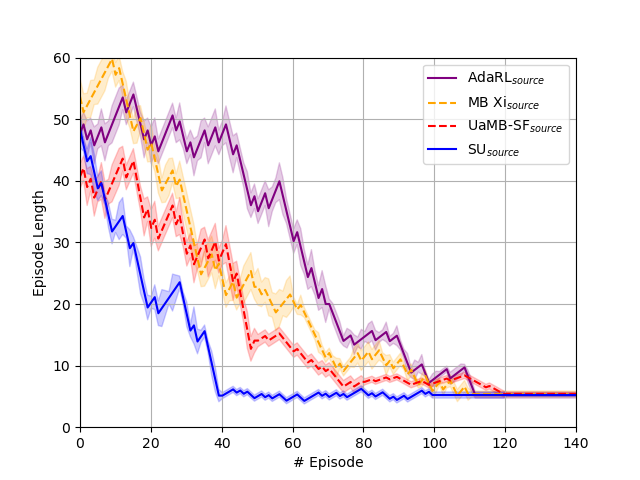}
}
\caption{ Episode length for source tasks A and 1. We present the average (solid lines) and
standard deviation (shaded area) of the episode length over $20$ runs generated from random seeds. The
shorter the episode, the sooner the agent reaches its goal. }
\label{fig:experiment_source}
\end{figure}
}
{ \begin{table}[tp]
\caption{Episode length averaged across all episodes for source tasks A and 1.}\label{Table:A}
\centering
\begin{tabular}{|P{1.5cm}|P{.8cm}|P{1.35cm}|P{1.3cm}|P{1.7cm}|}
 \hline
 \small{Source task} &  \small{$\text{SU}_{\text{source}}$} &   \small{$\text{AdaRL}_{\text{source}}$} & \small{$\text{MB Xi}_{\text{source}}$}  &  \small{$\SR_{\text{source}}$}
\\ \hline
\small{A} &  \small{ $\mathbf{50.8} \pm (5.9) $} &  \small{$91.9 \pm (5.0) $} &  \small{$71.6 \newline \pm (4.7) $} & \small{$66.1 \newline \pm (4.8)$}
 \\ \hline
\small{1} &   \small{$\mathbf{9.3} \pm (1.6) $} &  \small{$19.6 \pm (2.2)$} &  \small{$16.0 \pm (2.2) $} & \small{$13.3 \newline \pm (1.9)$}
\\ \hline
\end{tabular}
\end{table}
}
%
%

\subsection{Transfer learning}
In this sub-section, we compare the performance of $\SR$ with other methods in three transfer settings where the source and test tasks differ in reward functions, transition dynamics, and both reward functions and transition dynamics.

\subsubsection{Transfer with reward changes}
We examine if our proposed method can adapt its knowledge from source task A to improve sample efficiency while learning a test task with a different reward function than the source task.

\textbf{Test task:}
Task B, represented in Fig.~\ref{subfig:taskb}, is considered as the test task. It is generated by exchanging source task A's start and goal locations. With the preservation of feature vectors and transition dynamics, changing the goal location in a task is equivalent to changing the task's reward function.

\textbf{Experimental setup:}
All parameters of $\text{SU}_{\text{source}}$, $\text{MB Xi}_{\text{source}}$, $\text{AdaRL}_{\text{source}}$, and $\SR_{\text{source}}$ (i.e., $ \{\bm{\phi}(\s)\}_{\s \in \mS}, \{\F^a,\bm{\Pi}^a, \bt^a, \S^a \}_{a \in \mA}$, $\F^{\pi}$, and  $\bt^{\pi} $), which have been trained on the source task A, are transferred to the test task B to initialize parameters of $\text{SU}_{\text{test}}$, $\text{MB Xi}_{\text{test}}$, $\text{AdaRL}_{\text{test}}$, and $\SR_{\text{test}}$, respectively. $\text{SU}_{\text{test}}$, $\text{MB Xi}_{\text{test}}$, $\text{AdaRL}_{\text{test}}$, and $\SR_{\text{test}}$ will be then trained on the test task B over $500$ episodes, each with a maximum length of $200$ steps.

{ \textbf{Results: }
Fig.~\ref{subfig:resultb} compares episode length of $\text{SU}_{\text{test}}$, $\text{MB Xi}_{\text{test}}$, $\text{AdaRL}_{\text{test}}$, and $\SR_{\text{test}}$ averaged over $20$ repeats. As expected, transferring knowledge of SU, MB Xi, AdaRL, and $\SR$ across tasks with different rewards leads to faster convergence than learning from scratch, as in the source task A. This is because all SU, MB Xi, AdaRL, and $\SR$  algorithms learn the reward function from environmental samples.  As a result, the reward function is immediately updated when the received reward changes. This change in the reward function will then be propagated throughout the entire state space when it is incorporated into the value function.}
%
{
\begin{figure}[tp]
\centering
\subfigure[Test task B: differs from the source task A in the reward function. ]{
\label{subfig:taskb}
\includegraphics[scale=0.36]{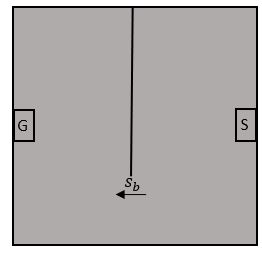}
}
\quad \quad \quad \quad \quad \quad 
 \subfigure[Episode length for task B.]{
\label{subfig:resultb}
\includegraphics[scale=0.34]{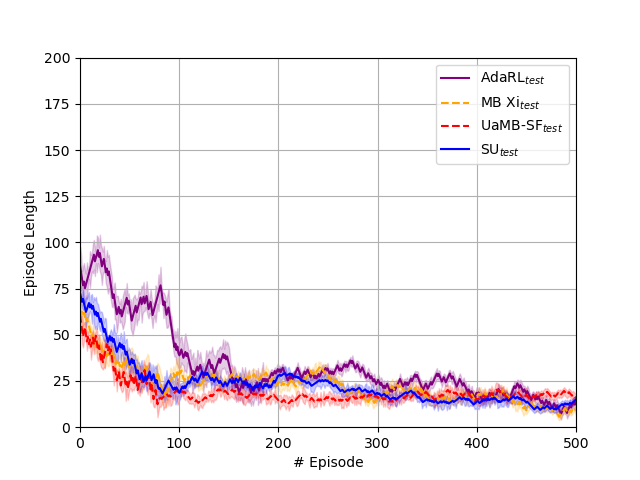}
}
\caption{ Test task B and its episode length averaged over $20$ runs.} \label{taskB}
\label{fig:taskB}%
\end{figure}
}
\subsubsection{Transfer with transition dynamics changes}
We now explore the knowledge generalization ability of SU, MB Xi, $\SR$, and AdaRL across tasks with various transition dynamics.

\textbf{Test task:} We consider task C, shown in Fig.~\ref{subfig:taskc}, to be the test task that has a different transition dynamics than the source task A. It is created by changing the position of the barrier in task A.

\textbf{Experimental setup:}
 The parameters of $\text{SU}_{\text{source}}$, $\text{MB Xi}_{\text{source}}$, $\text{AdaRL}_{\text{source}}$, and $\SR_{\text{source}}$ that were trained on the source task A are reused to initialize parameters of  $\text{SU}_{\text{test}}$, $\text{MB Xi}_{\text{test}}$, $\text{AdaRL}_{\text{test}}$, and $\SR_{\text{test}}$, respectively.  We then train $\text{SU}_{\text{test}}$, $\text{MB Xi}_{\text{test}}$, $\SR_{\text{test}}$, and $\text{AdaRL}_{\text{test}}$ on the test task C over $500$ episodes, each with a maximum length of $200$ steps.

{\textbf{Results:}
 Fig.~\ref{subfig:resultc} presents the results. It should be noted that any trial that does not complete the task in at most $22$ time steps fails to find the optimal policy.
Based on the results, we observe the inflexibility behaviour of SU and $\text{MB Xi}$ during the transfer across various transition dynamics. This is because SU and $\text{MB Xi}$ use the TD learning method to approximate the SF and the $\epsilon$-function,  which both store multi-step predictive maps rather than step-by-step transition dynamics. Hence, TD cannot update $\m^{\pi}(\s)$ and the $\epsilon$-function for states that are not directly affected by the change. When learning the test task C, the agent mimics the learned actions from the source task A until it encounters the barrier at $\s_b$. $\text{SU}_{\text{test}}$ and $\text{MB Xi}_{\text{test}}$ then update the SF and the $\epsilon$-function for state $\s_{b^-}$ that is immediately adjacent to $\s_b$, but the values of remaining states remain unchanged from the source task. Accordingly, the agent does not infer that states on the right side of $\s_b$ no longer follow states on the left side of $\s_b$ and thus chooses actions learned from the source task A (i.e., moving right) at these states and will end at state $\s_{b^-}$.
\\
However, knowledge transfer in $\SR$ and AdaRL results in learning task C using fewer samples than the source task A. This follows because both $\SR$ and AdaRL learn the model of transition dynamics using environmental samples and can thus update the model immediately after encountering a change. For instance, when the agent notices the barrier by being dropped at $\s_{b^-}$ rather than $\s_b$ after choosing to move right, $\SR_{\text{test}}$ immediately updates $\F^{a}_{\text{test}}$ (and thus $\F^{\pi}_{\text{test}}$) and $\S^{a}_{\text{test}}$ for $a=\text{'moving right'}$. The change in $\F^{\pi}$ is then distributed throughout the state space by $(\I-\gamma \F^{\pi})^{-1}$ in Eq.~\eqref{Eq:factor}. Additionally, as the results indicate, $\SR_{\text{test}}$  reaches the goal faster than $\text{AdaRL}_{\text{test}}$. This demonstrates the advantage of computing the value function as a single dot product in $\SR$ over the value iteration methods used in MB algorithms.}
{
\begin{figure}[bp]
\centering
\subfigure[Test task C: differs from the source task A in the transition dynamics. ]{
\label{subfig:taskc}
\includegraphics[scale=0.34]{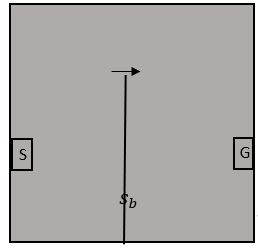}
}
\quad \quad \quad \quad \quad \quad 
 \subfigure[Episode length for task C.]{
\label{subfig:resultc}
\includegraphics[scale=0.34]{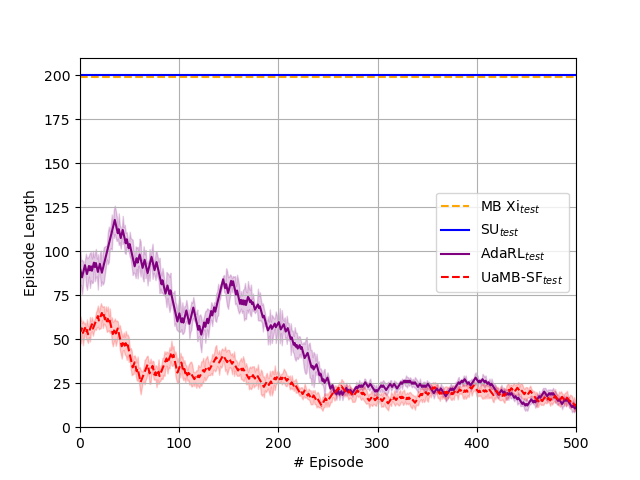}
}
\caption{ Test task C and its episode length averaged over $20$ runs.}
\label{fig:taskC}%
\end{figure}
}
\subsubsection{Transfer with reward and transition dynamics changes}
In this sub-section, we investigate whether $\SR$ can adapt its knowledge across different reward functions and transition dynamics.

\textbf{Test task:}   We consider the test task 2, shown in Fig.~\ref{subfig:2}, to be the test task. Task 2 differs from source task 1 in both the reward function and transition dynamics since it is generated by reversing the rotation direction of source task 1's left dial and changing the reward values so that setting the left dial to $2$ and the middle dial to $3$ is rewarding.

\textbf{Experimental setup:}
We transfer all parameters of $\text{SU}_{\text{source}}$, $\text{MB Xi}_{\text{source}}$, $\text{AdaRL}_{\text{source}}$, and $\SR_{\text{source}}$ that were trained on the source task 1 to task 2 to initialize $\text{SU}_{\text{test}}$, $\text{MB Xi}_{\text{test}}$, $\text{AdaRL}_{\text{test}}$, and $\SR_{\text{test}}$, respectively. $\text{SU}_{\text{test}}$, $\text{MB Xi}_{\text{test}}$, $\text{AdaRL}_{\text{test}}$, and $\SR_{\text{test}}$ are then trained on the test task 2 across $140$ episodes, each with a maximum length of $60$.

{\textbf{Results:} Fig.~\ref{subfig:steps2} plots the episode length for learning the test task 2 using the knowledge acquired from task 1. It is worth noting that any trial that did not complete the task in $5$ time steps failed to find the optimal policy. As the results show, in contrast to AdaRL and $\SR$, which afford positive knowledge transfer across various transition dynamics and rewards, TD learning-based algorithms such as $\text{SU}_{\text{test}}$ and  $\text{MB Xi}_{\text{test}}$ over-fit to a specific MDP. As a result, knowledge transfer in such algorithms has a negative impact because it prevents the agent from learning the optimal policy.}
%
%
 {
\begin{figure}[tp]
\centering
 \subfigure[Test Task 2: differs from source task 1 in both the reward and transition dynamics.]{
\label{subfig:2}
\includegraphics[scale=0.4]{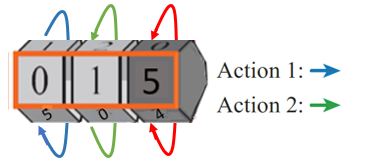}
}
\quad \quad \quad \quad \quad \quad 
 \subfigure[Episode length for task 2.]{
\label{subfig:steps2}
\includegraphics[scale=0.4]{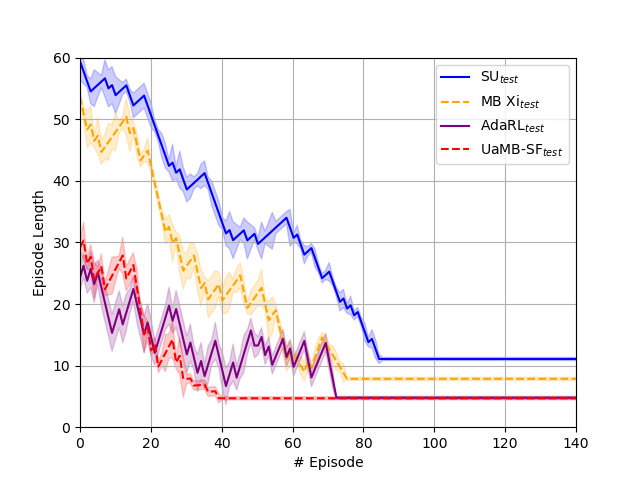}
}
\caption{ Test task 2 and its episode length averaged over $20$ runs.}
\label{fig:task2}
\end{figure}
}

{The results of all transfer learning experiments are summarized in Table~\ref{Table:tests}.}
 \afterpage{
\begin{table}[tp]
\caption{Episode length averaged across all episodes for the transfer learning experiments.}\label{Table:tests}
\centering
\begin{tabular}{|P{1.4cm}|P{.7cm}|P{1.1cm}|P{1.1cm}|P{1.7cm}|}
 \hline
 \small{Test task} &  \small{$\text{SU}_{\text{test}}$} &    \small{$\text{MB Xi}_{\text{test}}$} & \small{$\text{AdaRL}_{\text{test}}$} &   \small{$\SR_{\text{test}}$}
\\ \hline
\small{B} &  \small{ $23.2 \pm (2.9)  $} &  \small{$23.0 \pm (3.0) $} &  \small{$33.8 \pm (4.1) $} & \small{$\mathbf{19.4} \newline \pm (2.9)$}
 \\ \hline
\small{C} &  \small{ $200 \pm (0)  $} &  \small{$200 \pm (0) $} &  \small{$46.5 \pm (4.0) $} & \small{$\mathbf{26.6} \newline \pm (3.8)$}
 \\ \hline
\small{2} &  \small{ $21.9 \pm (1.8) $} &  \small{$15.3 \pm (1.3) $} &  \small{$8.2 \pm (0.9)$} & \small{$\mathbf{7.1} \newline \pm (0.9)$}
 \\ \hline
\end{tabular}
\end{table}
}
{\subsection{Ablation studies} \label{sec:ablation}
In this section, we conduct two sets of ablation studies to better understand the impacts of $\SR$'s two key components, the uncertainty-aware exploration scheme and the learned model (i.e., MB), on its performance.}
{\subsubsection{Effect of the uncertainty-aware exploration} \label{sec:ablation1}
The uncertainty-aware action selection model, which directs an agent to choose the most informative actions, is a key component of $\SR$. To evaluate how much the proposed uncertainty-aware exploration approach improves the learning process of a given task, we  compare $\SR$ to MB-SF$(\epsilon)$, a version of $\SR$ that learns the MDP model jointly with the feature vectors $\bm{\phi}$ by minimizing the loss objective $J\t$ in Eq.~\eqref{Eq:loss} using stochastic gradient descent. The agent in MB-SF$(\epsilon)$ does not capture uncertainty about the approximations and chooses actions using the $\epsilon$-greedy method. Each $\epsilon$-greedy policy selects actions uniformly at random with a probability of $\epsilon$ and otherwise chooses the action that maximizes the Q-value function.}

{ \textbf{Experimental setup:} We select the value of $\epsilon$ by performing a grid search for $\epsilon \in \{0.1,0.5, 0.2,0.3,0.4, 0.5, 0.6, 0.7, 0.8 \}$ and choosing the value that results in the fastest training. We then train MB-SF$(\epsilon)$ and $\SR$ on tasks A and 1 with the initial values provided in Table~\ref{Table:param_nav} of the appendix.}

{ \textbf{Results:}  The evaluation performance in terms of episode length is presented in Fig.~\ref{ablation}. The results confirm that the uncertainty-based exploration bonus significantly accelerates the learning of tasks A and 1, both of which have sparse reward functions and thus are challenging for exploration. To further illustrate the effect of the proposed uncertainty-aware exploration, error bounds on the approximated Q-value function in Eq.~\eqref{Eq:error_bound} for MB-SF$(\epsilon)$ and $\SR$ are plotted in Fig.~\ref{fig:loss_A}. As shown, the proposed exploration scheme helps $\SR$ to minimize its approximation error faster than MB-SF($\epsilon$) and thus makes the convergence faster.
}
 {
\begin{figure}[bp]
\centering
\subfigure[Episode length for task A. ]{
\label{subfig:taskA_ablation}
\includegraphics[scale=0.32]{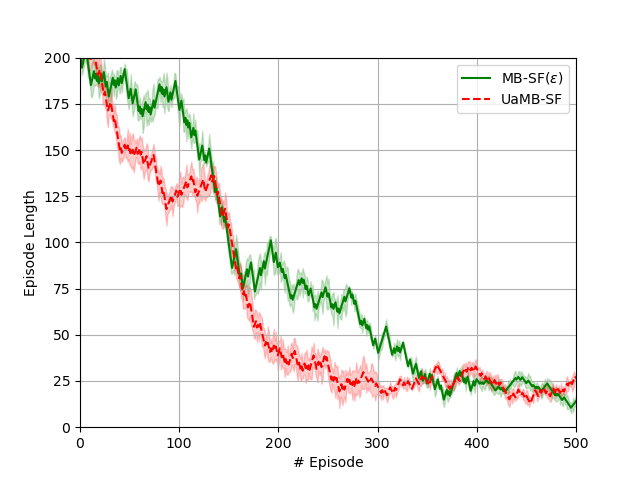}
}
\quad \quad \quad \quad \quad 
 \subfigure[Episode length for task 1.]{
\label{subfig:task1_ablation}
\includegraphics[scale=0.32]{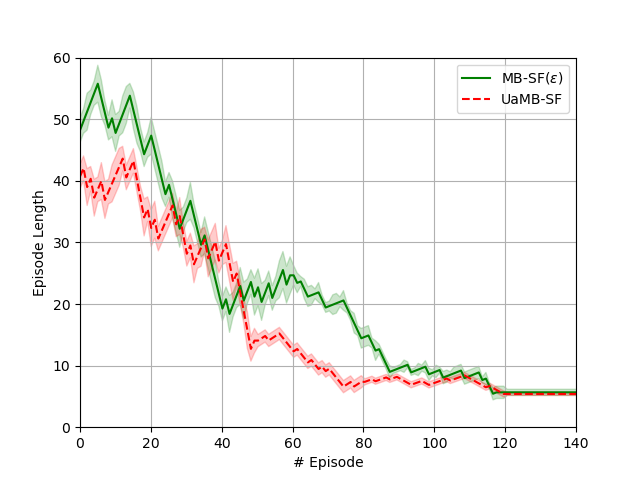}
}
\caption{ Ablation results for investigating the effect of the uncertainty-aware exploration in performance of $\SR$.}
\label{ablation}
\end{figure}
\begin{figure}[tp]
\centering
\subfigure[Error bound on the approximated Q-value function for task A. ] {\includegraphics[scale=0.31]{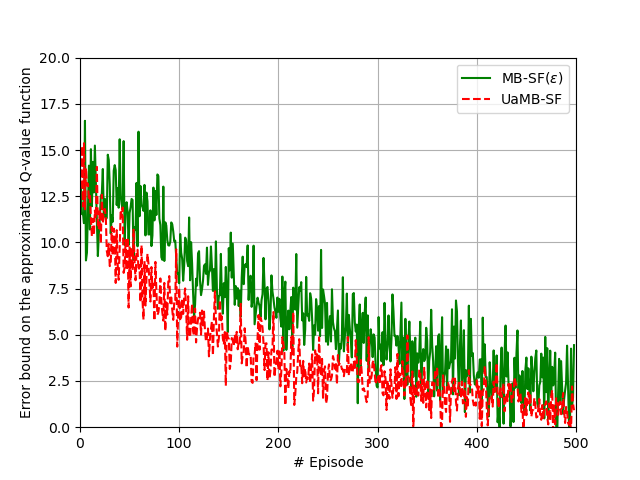}}
\quad \quad \quad \quad \quad  
\subfigure[Error bound on the approximated Q-value function for task 1.  ] {\includegraphics[scale=0.31]{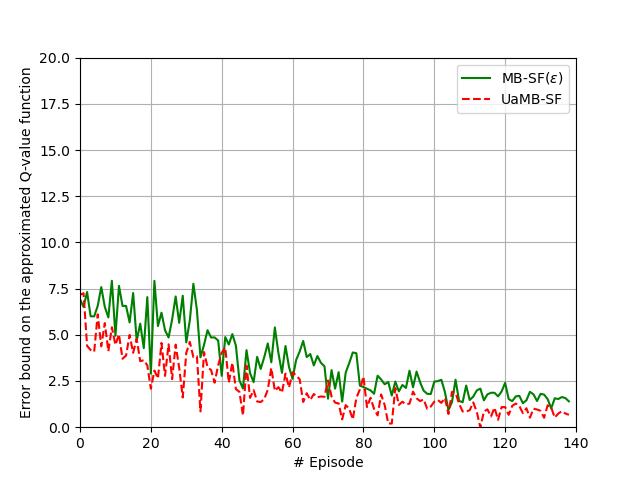}}
\caption{ Ablation results for investigating the effect of the uncertainty-aware exploration in performance of $\SR$.}
 \label{fig:loss_A}
\end{figure}
}
{\subsubsection{Effect of the learned model} \label{sec:ablation2}
To demonstrate how learning the model of transition dynamics (i.e., the MB component) in $\SR$ enables the agent to adapt its knowledge across different transition dynamics, we compare the performance of $\SR$ to that of UaTD-SF, a variant of $\SR$ that does not learn transition dynamics of its environment. UaTD-SF learns the SF using KF-based multiple-model adaptive estimation, with the measurement model derived from the TD equation in Eq.~\eqref{Eq:SF-TD22} as $\bm{\phi}\t(\s)=\m\t^{\pi}(\s)-\gamma \m\t^{\pi}(\s')$. Similarly to $\SR$, UaTD-SF leverages the uncertainty information about $\m\t^{\pi}$ to select the actions that lead to the greatest reduction in the uncertainty of the Q-value function.}

{ \textbf{Experimental setup:} Consider $\text{UaTD-SF}_{\text{source}}$ as the implementation of UaTD-SF on the source task A. After training $\text{UaTD-SF}_{\text{source}}$ and $\SR_{\text{source}}$  on task A from scratch over $500$ episodes, all of their trained parameters are reused to initialize $\SR_{\text{test}}$ and $\text{UaTD-SF}_{\text{test}}$, which will then be trained on the test task C over $500$ episodes.}

{ \textbf{Results:} Fig.~\ref{fig:ablation_MB} illustrates the results. While UaTD-SF learns the source task A faster than $\SR$, it cannot generalize its knowledge to test task C, which has different transition dynamics than task A. This is explained with the same reason provided in the motivating exam (Sub-section~\ref{sec:moti_exam}). UaTD-SF and other TD-SF methods that do not learn transition dynamics cannot update the SF value transferred from the source task for all states in the test task. This inflexibility of UaTD-SF results in negative knowledge transfer, preventing the agent from finding the optimal policy for the test task. However, learning transition dynamics in $\SR$ using environmental samples enables the agent to adjust the SF value for all states in the test task upon encountering a change in the transition dynamics. The updated value of the SF then causes the agent to adapt its actions from the source task to the test task.
}
 {
\begin{figure}[bp]
  \captionsetup{singlelinecheck = false}
\centering
\subfigure[Episode length for task A. ]{
\label{subfig:taskA_MB_ablation}
\includegraphics[width=0.32\textwidth]{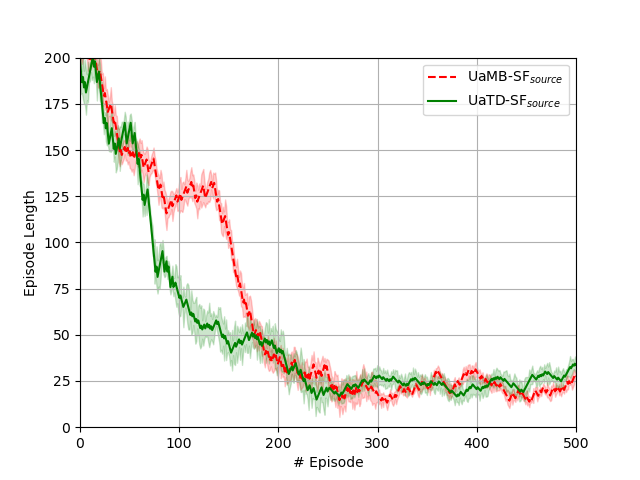}
}
\quad \quad \quad \quad 
 \subfigure[Episode length for task C.]{
\label{subfig:taskC_MB_ablation}
\includegraphics[width=0.32\textwidth]{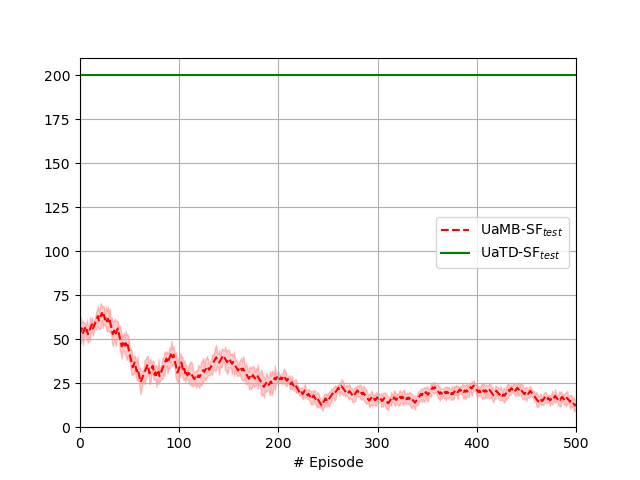}
}\\
\noindent \caption{ Ablation results for showing the effect of learning transition dynamics (i.e., the MB component) on $\SR$'s knowledge generalization ability.} \label{fig:ablation_MB}
\end{figure}
{\subsection{Sensitivity analysis studies}
Given the lack of information about the $\SR$'s parameters, determining how sensitive $\SR$ is to variations in such uncertain parameters is critical. These parameters, particularly the dimension of the state feature vector $\bm{\phi}(\s)$ and the measurement noise variance $P^N$ of the KF learning the reward function, are critical but poorly defined. As a result, we run two sensitivity analysis experiments in this sub-section to see how different values for $\bm{\phi}(\s)$'s dimension and $P^N$ affect $\SR$ overall performance. }
{\subsubsection{Effect of the feature vector dimension}
As the reward function and transition dynamics in $\SR$ are both functions of the $L$-dimensional feature vector $\bm{\phi}(\s)$, we now study the effect of varying the dimension $L$  (i.e., the number of RBFs) on $\SR$ performance.}

{\textbf{Experimental setup:}  We train $\SR_{\text{source}}$ on tasks A and 1 with different values of $L$ ranging from $L=9$ to $L=36$.}

{\textbf{Results:} The performance of $\SR$ for various values of $L$ is shown in Fig.~\ref{fig:effect_L} and Table~\ref{Table:L}.  According to the findings, $16$ RBFs are sufficient to learn an optimal policy for task A, and increasing the number of RBFs to $L=25$ results in a marginal improvement. However, we observe a drop in the performance for $L=16$ in task 1, preventing  the agent from learning an optimal policy. This is because the approximation of the reward function $R(\s,a) \approx \phi^T(\s)  \bt\t^a$ with a low number of RBFs induces a spatial smoothing~\cite{wu2018laplacian}, whereas the reward function of task 1 is non-zero at only one state. Therefore, a larger $L$ (such as $25$) is required to maintain accurate information about  task 1's reward function.}
{
\begin{figure}[bp]
\centering
\subfigure[Episode length of $\SR$ on task A. ]{
\label{subfig:taskA_L}
\includegraphics[scale=0.31]{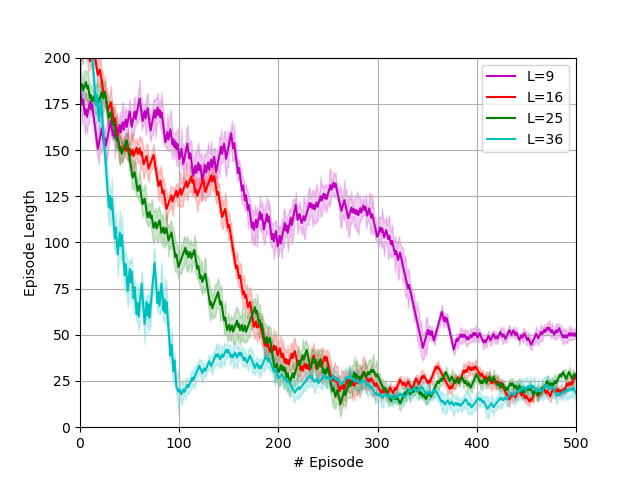}
} 
\quad \quad \quad 
 \subfigure[Episode length of $\SR$ on task 1.]{
\label{subfig:task1_L}
\includegraphics[scale=0.31]{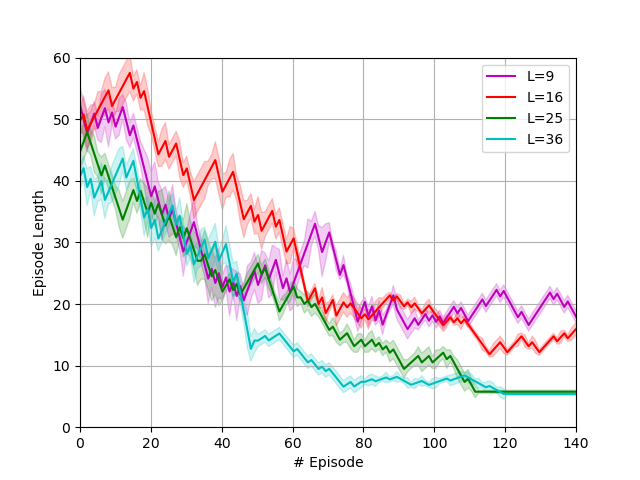}
}
  \caption{ {Performance of $\SR$ for different dimensions $L$.}}
  \label{fig:effect_L}%
\end{figure} 
}
%
\afterpage{
\begin{table}[tp]
\caption{{Episode length of $\SR$ averaged across all episodes for different dimensions $L$. }}\label{Table:L}
\centering
\begin{tabular}{|P{.8cm}|P{.9cm}|P{1.cm}|P{1.cm}|P{1.1cm}|}
 \hline
 \small{ Task} &  \small{$L=9$} &    \small{$L=16$} & \small{$L=25$} &   \small{$L=36 $}
\\ \hline
\small{ A} &  \small{ $106 \pm (5.8)  $} &  \small{$66.1 \pm (4.8) $} &  \small{$56.2 \pm (5.0) $} & \small{$\mathbf{39.5} \pm (4.8)$}
 \\ \hline
\small{1} &  \small{ $20.3 \pm (1.7)  $} &  \small{$25.3 \pm (1.3)$} &  \small{$15.3 \pm (1.1) $} & \small{$\mathbf{13.3} \pm (1.0)$}
 \\ \hline
\end{tabular}
\end{table}
}
{\subsubsection{Effect of the measurement noise variance $P^{N}$} \label{sec:abl_noise}
One of the most important parameters for determining the performance of a KF is its measurement (co)variance, which is a design parameter. This (co)variance can be chosen if some knowledge of the measurement model is available.  However, such information is generally difficult to obtain in advance. In this work, the measurement noise (co)variances are chosen by trial and error.  This sub-section investigates how different values for the measurement noise variance $P^N$ in the reward learning module affect $\SR$ performance.}

{\textbf{Experimental setup:}  We train $\SR$ on tasks A and 1 with three different $P^N=0.01, 0.1, 1$ values while keeping the other parameters constant.}

{\textbf{Results:} The results are depicted in Fig.~\ref{fig:effect_covar}. The best results for tasks A and 1 are obtained when $P^N=0.1$ and $P^N=1$, respectively. In both tasks, $\SR$ with $P^{N}=0.01$ converges to the final goal much more slowly, necessitating more episodes to learn the tasks.  Several studies, such as~\cite{park2019measurement, wang2018kalman}, have been conducted to estimate parameter values of a KF. We leave the investigation of estimating these parameters for future work.}
%
\afterpage{
\begin{figure}[tp]
\centering
\subfigure[Episode length of $\SR$ on task A. ]{
\label{subfig:taskA_covar}
\includegraphics[scale=0.31]{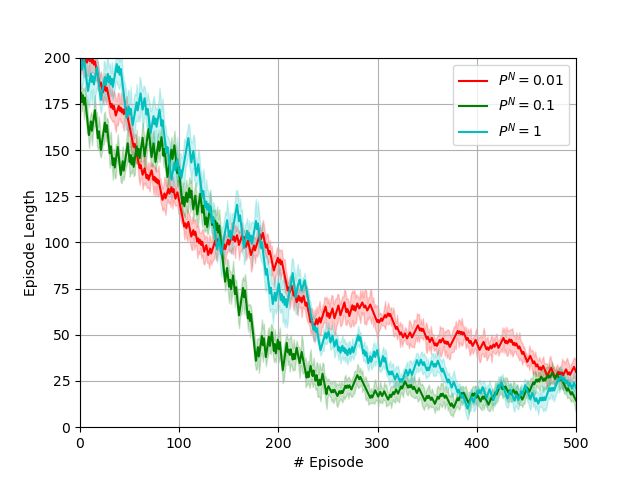}
} 
\quad \quad \quad 
 \subfigure[Episode length of $\SR$ on task 1.]{
\label{subfig:task1_covar}
\includegraphics[scale=0.31]{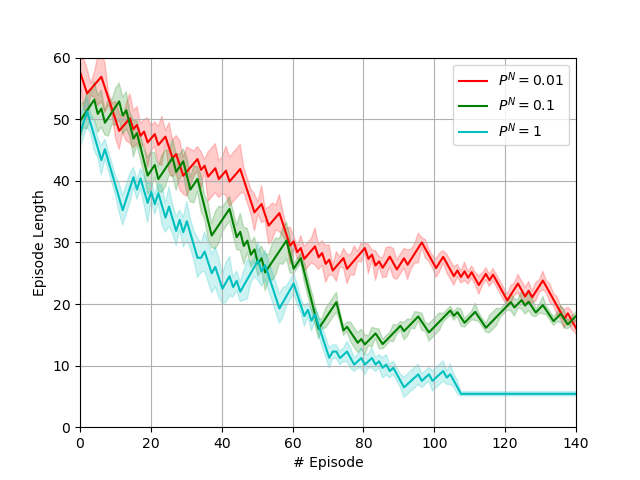}
}
\caption{ {Performance of $\SR$ for different values of the measurement noise variance $P^N$.}  }
\label{fig:effect_covar}%
\end{figure}
}
 \section{Algorithmic complexity  \label{sec:complexity}}
To calculate the global computational complexity of the proposed $\SR$ scheme, we  deduce the complexity of each of its modules as follows: (i) learning the reward function $R$ using a KF-based multiple-model adaptive estimation with $M_{\text{KF}}=1 $ necessitates inverting the scalar value $\big(\h\t \bm{\Pi}^a_{t|t-1}\h^T\t +P^N\t \big) $,  (ii) learning the transition dynamics through a matrix-based KF requires inverting a ($L \times L$) matrix, (iii) learning state features requires inverting the ($D \times D$) matrix $\{\bm{\Sigma}\}^{(i)}_{i=1:L}$, where $D$ is the dimension of each state $\s \in \mS$, (iv) learning the SF needs inverting the ($L \times L$) matrix $\left(\I-\gamma\bm{F}^{\pi}\t \right)^{-1}$, and (v) action selection via the proposed uncertainty-aware exploration requires trace calculation of a ($D \times D$) matrix.
The proposed $\SR$ algorithm, therefore, has a global computational complexity (per iteration) of  $O\, ( 2 L^3 + L \times D^2)$, where usually $L >> D$.
\\ {It should be noted that the primary focus of this paper is on improving the sample efficiency of RL agents via knowledge transfer across various rewards and transition dynamics, as well as uncertainty-oriented exploration.  It is expected that TD-based algorithms such as SU~\cite{SU} and MB Xi~\cite{reinke2021xi} have less complexity than $\SR$ and AdaRL~\cite{huang2021adarl} since they do not learn transition dynamics. However, TD-SF methods are limited to knowledge generalization across different rewards only.  Moreover, $\SR$ has lower computational complexity than AdaRL. This is because, given the MDP model, AdaRL computes the value function using value iteration algorithms that require intractable integrals in large and continuous state spaces, whereas $\SR$ computes the value function using a dot product of the SF and the reward parameter $\bt^a$.}
{\section{ Discussion and future extensions} \label{sec:disc} 
This paper presents $\SR$, a novel approach for knowledge adaptation across different reward functions and transition dynamics in RL domains. This method is inspired by combining aspects of MB and SF algorithms and ties SF learning to MB RL. Although this link between MB and SF methods has previously been hypothesized~\cite{momennejad2017, tomov2021multi}, $\SR$ formalizes it. $\SR$ generalizes its knowledge to variations in both reward functions and transition dynamics. This ability distinguishes $\SR$ from previous SF-based transfer learning approaches such as SU~\cite{SU} and MB Xi~\cite{reinke2021xi}, which only demonstrate adaptability against changes in reward function. Moreover, $\SR$ is less computationally demanding than previous MB work, such as AdaRL~\cite{huang2021adarl}. Furthermore, unlike MB methods, $\SR$ models the reward function and transition dynamics parameters as random vectors, allowing us to compute uncertainties of the parameters.  $\SR$ leverages estimated uncertainties to derive a type of uncertainty-oriented exploration. The ablation experiments presented in Sub-section~\ref{sec:ablation} demonstrate the importance of incorporating uncertainty into the action selection process in improving sample efficiency while learning a task. Additionally, contrary to current MB methods, the proposed $\SR$ framework can deal with the possible non-stationarity of an MDP model in real-world problems.}

{ While $\SR$ can efficiently adapt its knowledge from source tasks to complete target tasks using fewer samples than learning from scratch, important limitations remain. One open question central to model learning methods and thus $\SR$ is how to handle approximation errors in the reward function and transition dynamics. The approximation bound presented in Eq.~\ref{Eq:error_bound} shows how the discount factor $\gamma$ influences the accuracy of the learned Q-value function. Specifically, the accuracy of the approximated Q-value function after $T$ time steps into the future is discounted by a factor of $(\frac{1+\gamma}{1-\gamma})^T$. Hence, the approximation error will generally grow as $T$ increases, and $\gamma$ tends to one. Consequently, the following questions arise: how large should $\gamma$ be, and how many transition samples are required to accurately approximate the Q-value function versus  the reward function and transition dynamics? Future research would explore the design of $\gamma$ and compare learning the Q-value function to learning the reward function and transition dynamics.}
\\
{ Moreover, it would be interesting to explore problems with higher levels of novelty between source and target tasks. The state and action space (thus, the state feature space) between the source and target tasks were assumed to be constant in $\SR$. However, as demonstrated in Section~\ref{sec:appendix_additional} of the appendix, $\SR$ cannot generalize its knowledge if the state feature vectors learned in the source task cannot be reused in the test tasks. An interesting perspective is to transfer knowledge across tasks with different action and state spaces (thus, different feature vectors).\\
Furthermore, multi-agent RL~\cite{Mohammad:Icassp, zhang2020model} is an essential sub-field of RL. Extending existing transfer learning techniques to the multi-agent scenario is challenging because multi-agent RL typically necessitates socially desirable behaviors~\cite{ndousse2021emergent}. To our knowledge,  \citet{meng2021offline} represents the only effort in multi-agent knowledge generalization across different reward functions. It remains unclear how to extend $\SR$'s knowledge adaptation ability across different transition dynamics to multi-agent RL, which we believe is a promising research direction.
\\
In addition, as shown in Sub-section~\ref{sec:abl_noise}, we could empirically improve the performance of $\SR$ by properly designing KF's parameters, particularly noise covariances, which are problem-dependent. It should be noted that choosing values for these parameters is not more difficult than choosing learning rates commonly used in the RL community; it is just less familiar. Several studies, such as~\cite{park2019measurement, wang2018kalman}, have been conducted to estimate parameter values of a KF. We leave the estimation of KF's parameters for future work.
\\
Eventually, our experiments have been limited to tasks with discrete action spaces. We hope that this work can be extended to tasks with continuous action spaces.}
\section{Conclusion}
\label{sec:con}
In this paper, we studied the problem setting of transfer learning in RL: a situation in which an agent must leverage prior experience from a trained task to complete a new task with less computation and fewer samples. We started by explaining why current SF-based transfer learning algorithms, which learn the SF with TD learning, can only generalize and adapt their knowledge across tasks with different reward functions. Motivated by the efficient computation of SF methods and the adaptability of MB algorithms in response to changes in rewards and transition dynamics, we then presented a hybrid MB-SF transfer learning algorithm. MB-SF enables the agent to generalize its knowledge across variations in both transition dynamics and reward function and contrasts with the computationally expensive nature of MB methods at decision time. The proposed MB-SF  provides the groundwork for the research workflow required for large-scale tasks in practical RL where previous computational work is available. MB-SF can also be used to demonstrate the computationally efficient flexible behaviour observed in the empirical literature. Finally, to improve sample efficiency even further, we proposed a novel uncertainty-oriented exploration strategy based on a KF-based multiple-model adaptive estimation that approximates the environment model, i.e., the MB component of MB-SF. The multiple-model estimation approach can also handle non-stationary and nonlinear approximations, allowing us to solve real-world problems that necessitate scaling up.
Our experiments on two RL problems verified that the hybrid MB-SF algorithm and the uncertainty-aware exploration enable the agent to adapt and better use its knowledge across various transition dynamics or/and reward functions. 
%
%
\clearpage

\begin{appendices}

\counterwithin{figure}{section}
\counterwithin{algorithm}{section}
\counterwithin{table}{section}
\counterwithin{equation}{section}
\section{Algorithm and theoretical results} \label{sec:Algorithm}
The proposed $\SR$ framework is briefed in Algorithm~\ref{algo:summary}.
\afterpage{
\begin{algorithm}[t]
\caption{\textproc{ $\SR$}}
\label{algo:summary}
\begin{algorithmic}[1]
\State {  \multiline{ \textbf{Input:}  \\ $ N_{\text{episode}}$: number of episodes \\ $\gamma$: discount factor  \\ $P^N$: measurement noise variance for reward learning \\ $\P^{\omega}$: process noise covariance for reward learning  \\ $\Sig^{\A}$: process noise covariance for transition dynamics learning \\ $\Sig^{\B} $: measurement noise variance for transition dynamics learning} }
\State {  \multiline{ \textbf{Initialize:} \\ $\{ \bt_0^a \}_{a \in \mA}$: parameter vector of the reward function \\ $\{ \F_{0}^a\}_{a \in \mA}$: parameter matrix of the transition dynamics \\ $\{\S_0^a\}_{a \in \mA}$: posterior covariance of $\{ \F_{0}^a\}_{a \in \mA}$ \\ $\{ \bm{\Pi}_{0}^a\}_{a \in \mA}$: posterior covariance of $\{ \bt_0^a \}_{a \in \mA}$ \\ $\{\bm{\phi}_0(\s)\}_{\s \in \mS}$: feature vector } }
\For{$n=1,2,..., N_{\text{episode}}$}
\State  $\s \leftarrow$ initial state
\For{$t=1,2, ...$}
\State \multiline{ {\textit{\textbf{Uncertainty-aware exploration:}}  \myindent{0} $a = \arg\max_{b\in\mA} [Q\pt(\s,b)$ \myindent{0} $+  \text{tr}\{\bm{\Pi}^b\pt+\S^b\pt\} ]$.}}
\State { Take action $a$ , observe $\s'$ and $ R\t(\s,a)$.}
\State { $\bm{\phi}\t(\s) \leftarrow \bm{\phi}\pt(\s)$ .}
\State \multiline{ {\textit{\textbf{Reward learning:}}
 Perform a KF to estimate $\bt^a\t$ and $\bm{\Pi}\t^a$.}}
 \State $\bt\t^{\pi} \leftarrow \bt\t^a$
\State \multiline{ \textit{\textbf{Transition dynamics learning:}} Perform a matrix-based KF to estimate $\F^a\t$ and $\S\t^a$.}
\State $\bm{F}^{\pi}\t \leftarrow\F^a\t$
\State Calculate $\text{tr}\{\bm{\Pi}^a\t+\S\t^a\}$.
\State  \multiline{\textit{\textbf{ State feature learning}}: Perform stochastic gradient descent on $J\t$ to update $\bm{\phi}\t(\s)$ }
\State  \multiline{ Compute the Q-value function: $Q\t(\s,a)=\left( (\bt^a\t)^T + \gamma (\bt^{\pi}\t)^T (1-\gamma \bm{F}^{\pi}\t)^{-1} \bm{F}^a\t\ \right) \bm{\phi}\t(\s)$.}
\State $\s \leftarrow \s'$
\EndFor
\EndFor
\end{algorithmic}
\end{algorithm}
}
\subsection{Appendix for Sub-section~\ref{sec:SR-MB11}} \label{sec:invert_A}
As illustrated in the main paper, the SR matrix $\M^{\pi}$ for finite states and actions can be computed as: $\M^{\pi}=(\I - \gamma \T^{\pi})^{-1}$, where $\T^{\pi}$ represents one step transition matrix such that $\bm{T}^{\pi}(\s,\s') = \sum_{a \in \mA} \pi(a|\s) P(\s'|\s,a)$. Now we prove that inverse of the matrix $(\I - \gamma \T^{\pi})$ exists:
 \vspace{.1in} \\
\textit{Proof}: Since $\T^{\pi}$ is a stochastic matrix, all its eigenvalues $\lambda_i \leq 1$.
The matrix ($\I - \gamma \T^{\pi}$) is thus invertible because
{ \begin{eqnarray}
\!\!\!\! \!\!\!\!  \text{det} (\I - \gamma \T^{\pi}) \geq \text{det} (\I) + (- \gamma)^{|\mS|} \text{det} (\T^{\pi})\nonumber \\
   \geq 1-\gamma^{|\mS|} \text{det} (\T^{\pi})=1- \gamma^{|\mS|} \prod_{i=1} \lambda_i > 0,
\end{eqnarray}}\normalsize
where det(.) depicts the determinant operator of a matrix.
\subsection{Appendix for Sub-section~\ref{sec:error_bound}}
\label{sec:error_A}
In this section, we formally prove that the error of the Q-value function approximation scales linearly in reward prediction error $e^R\t=\sup_{\s \in \mS, a \in \mA} |R\t(\s,a)-(\bt^a\t)^T \bm{\phi}\t(\s)|$ and transition approximation error $e^P\t=\sup_{\s \in \mS, a \in \mA} ||\mathbb{E}_{ P}[\bm{\phi}\t(\s\nt)]-\F\t^a \bm{\phi}\t(\s)||$:
{
\begin{eqnarray}
|Q\t^{\pi}(\s,a)-(\q^a\t)^T \bm{\phi}\t(\s)| \leq \frac{e^R\t+\gamma ||\v^{\pi}\t|| e^P\t}{1-\gamma}.
\end{eqnarray}}\normalsize
\textit{Proof}:
{ \begin{eqnarray}
\!\!\!\!\!\!\!\!\!\!\!\!\!\!\! |Q^{\pi}\t(\s,a)-(\q^a)\t^T \bm{\phi}\t(\s)| &\leq& \big|R\t(\s,a) + \gamma \mathbb{E} _{P} \left[V^{\pi}\t(\s')| \s,a \right]  -(\bt^a)\t^T \bm{\phi}\t(\s)- \gamma (\v^{\pi}\t)^T \bm{F}\t^a \bm{\phi}\t(\s)\big| \nonumber \\
 &\leq & |R\t(\s,a)-(\bt^a\t)^T \bm{\phi}\t(\s) | + \gamma \left| \mathbb{E} _{ P} \left[V\t^{\pi}(\s')|\s,a \right] - (\v^{\pi}\t)^T \bm{F}\t^a \bm{\phi}\t(\s\t)\right|.
 \label{Eq:Q-function}
\end{eqnarray}} \normalsize
The second term in Eq.~\eqref{Eq:Q-function} is bounded by
\begin{eqnarray}
\left|\mathbb{E}_{P} \left[V^{\pi}\t(\s')|\s,a \right] - (\v^{\pi}\t)^T \bm{F}^a\t \bm{\phi}\t(\s)\right| &=& \big|\mathbb{E} _{P} \left[V^{\pi}\t(\s')|\s,a\right] - \mathbb{E}_{P}[(\v^{\pi}\t)^T \bm{\phi}\t(\s\nt)|\s,a]   \nonumber \\
&+& \mathbb{E}_{P}[(\v^{\pi}\t)^T \bm{\phi}\t(\s\nt)|\s,a]  - (\v^{\pi}\t)^T \bm{F}^a\t \bm{\phi}\t(\s)\big| \\
 &=& \sup_{\s,a} \left|Q^{\pi}\t(\s,a)-(\q^{a}\t)^T \bm{\phi}\t(\s)\right| + \left|\mathbb{E}_{P}[(\v^{\pi}\t)^T \bm{\phi}\t(\s\nt)|\s,a] -(\v^{\pi}\t)^T \bm{F}^a\t \bm{\phi}\t(\s)\right| \nonumber\\
&=& \sup_{\s,a} \left|Q^{\pi}\t(\s,a)-(\q^{a}\t)^T\bm{\phi}\t(\s)\right|  +||(\v^{\pi}\t)|| \left|\mathbb{E}_{P}[ \bm{\phi}\t(\s\nt)|\s,a] - \bm{F}^a\t \bm{\phi}\t(\s)\right| \nonumber \\
&=&\sup_{\s,a} \left|Q^{\pi}\t(\s,a)-(\q^{a}\t)^T \bm{\phi}\t(\s)\right| + ||\v^{\pi}\t|| e^P\t.  \label{Eq:et}
\end{eqnarray}
Replacing Eq.~\eqref{Eq:et} into Eq.~\eqref{Eq:Q-function} results in
{ \begin{eqnarray}
\lefteqn{\!\!\!\!\!\!\!\!\!\!\!\!\!\!\!\! \forall \s,a:  \big|Q^{\pi}\t(\s,a)-(\q^a\t)^T \bm{\phi}\t(\s)\big| \leq \big|R\t(\s,a)-(\bt^a\t)^T \bm{\phi}\t(\s) \big|} \nonumber\\
&&\!\!\!\!\!\!\!\!\!\!\!\!\!\!\!\!\!\!\!\! + \gamma \left( \underbrace{\sup_{\s,a} \left|Q^{\pi}\t(\s,a)-(\q^a\t)^T \right|}_{E\t} + ||(\v^{\pi}\t)^T|| e^P\t \right).
\
\end{eqnarray}} \normalsize
Therefore:
{ \begin{eqnarray}
E\t \leq e\t^R + \gamma E\t + ||(\v^{\pi}\t)^T || e^P\t,
\end{eqnarray}} \normalsize
which implies that
{ \begin{eqnarray}
E\t  \leq \frac{e^R\t+\gamma ||\v^{\pi}\t|| e^P\t}{1-\gamma}.
\end{eqnarray}} \normalsize
\subsection{ Kalman filter  \label{sec:KF}}
In this section, we briefly explain the Kalman filter (KF)~\cite{KF}, which is the foundation of the formulation presented in this work. The KF is used for learning parameters $\x$ of a linear approximation function with the next evolution and measurement (update) models:
\begin{align}
& \text{Evolution model:} \hspace{1cm} \x\t = \G\t \x\pt + \bm{v}\t,  \\
 & \text{Measurement model:} \hspace{.8cm}  \y\t = \H\t \x\t  + \bm{n}\t ,
\end{align}
where $\x\t$ and $\y\t$ are the parameter and observation scalars or vectors at the current time step $t$, respectively. $\G\t$ is the evolution function and $\H\t$ represents measurement mapping function. $\v\t$ is the evolution noise, and $\n\t$ is the measurement noise,
both modelled as zero-mean white Gaussian  noises with covariances $\P^{\v}\t$ and $\P^{\n}\t$, respectively. KF treats the parameter $\x\t$ as a random variable. $\hat{\x}_{t|t}=\mathbb{E}[\x\t|\y_{1:t}]$ represents a posteriori estimation of $\x$ at time $t$ given observations up to and including at time $t$ (i.e., $\y_{1:t}$). Covariance (uncertainty) of the posteriori estimate is denoted by $\P_{t|t}=\mathbb{E}[(\x\t-\hat{\x}_{t|t})(\x\t-\hat{\x}_{t|t})^T|\y_{1:t}] $. $\hat{\x}_{t|t}$ and $P_{t|t}$ are obtained through two distinct phases: predict and update. The predict phase does not include observation information from the current time step $t$ and uses the state estimate from the previous time step to produce an estimate of the state at the current time step:
\small
\begin{align}
&\!\!\!\!\!\! \!\!\!\!\!\!\!\! \!\!  \textbf{Predict}  \nonumber \\
& \!\!\!\!\!\!\!\!\!\!\!\!\!\!\!\!    \text{  Predicted state estimate:} \hspace{.4cm} \hat{\x}_{t|t-1} = \G\t \hat{\x}_{t|t-1}, \\
& \!\!\!\!\!\!\!\!\!\!\!\!\!\!    \text{Predicted estimate covariance:} \hspace{.1cm}  \bm{P}_{t|t-1} = \G\t \bm{P}_{t-1|t-1} \G\t^T+ \P^{\v}\t.
\end{align}
 \normalsize
 In the update phase, the difference between the current a priori prediction ($\H\t  \hat{\x}_{t|t-1}$) and the current observation ($\y\t$) is multiplied by the Kalman gain $K\t$ and combined with the previous state estimate to refine the state estimate. This improved estimate based on the current observation is called the posteriori state estimate:
\small
\begin{align}
& \!\!\!\!\!\!\!\!\!\!\!\!\!\!\!\!\!\!\! \textbf{ Update} \nonumber  \\
& \!\!\!\!\!\!\!\!\!\!\!\!\!\!\!\!\!\! \text{Kalman gain:}  \hspace{.8cm} \K\t^a = \bm{P}_{t|t-1} \H^T\t \big(\H\t \bm{P}_{t|t-1}\H\t^T +\P^{\n}\t \big)^{-1}, \\
& \!\!\!\!\!\!\!\!\!\!\!\!\!\!\!\!\!\! \text{ Posteriori state estimate:} \hspace{.1cm} \hat{\x}_{t|t} = \hat{\x}_{t|t-1} + \K\t\big(\y\t - \H\t  \hat{\x}_{t|t-1} \big), \\
& \!\!\!\!\!\!\!\!\!\!\!\!\!\!\!\!\!\! \text{  Posteriori estimate covariance: } \hspace{.2cm} \bm{P}_{t|t} = \big(\I - \K\t \H\t\big)\bm{P}_{t|t-1} .
\end{align}
  \normalsize
 \\
 Once the parameter $\hat{\x}_{t|t}$ and its uncertainty ($\bm{P}_{t|t}$) are estimated, the likelihood $\text{Pr}(\y\t|\x\t)$ and posterior distribution $\text{Pr}(\x\t|\y\t)$ are Gaussian as
\begin{eqnarray}
 \y\t &\sim & \mathcal{N}(\H\t \hat{\x}_{t|t},\P^{\n}\t), \\
 \x\t &\sim & \mathcal{N}( \hat{\x}_{t|t},\bm{P}_{t|t}).
\end{eqnarray}
Fig.~\ref{fig:KF} represents the KF perspective for estimating a three-dimensional parameter vector $\x$.
\begin{figure}[t]
\centering
\includegraphics[scale=.44]{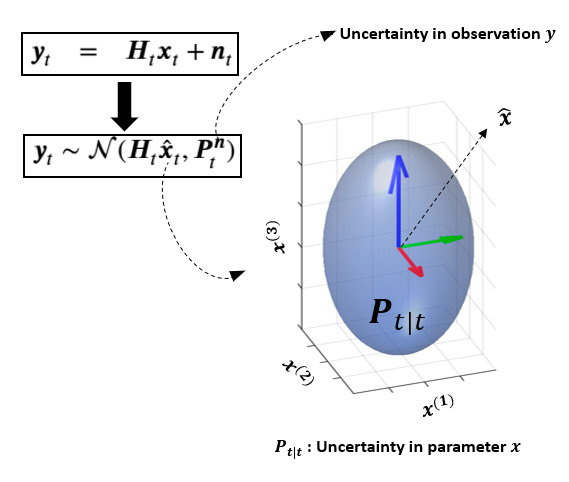}
\caption{ Kalman filter perspective for estimation of three-dimensional parameter vector $\x\t$ from observation $\y\t$ at time step $t$. $\bm{P}_{t|t}$ is the epistemic uncertainty in the parameter $\x\t$.}
\label{fig:KF}
\end{figure}
In this article, the tilde over the estimated parameter was omitted for ease of notation.
\subsection{ multiple-model adaptive estimation \label{sec:MMAE}}
If we represent the $i^{\text{th}}$ KF of the whole $M_{\text{KF}}$ KFs with $m\i$, the weight of $m\i$ is calculated recursively using the Bayesian rule as
\begin{eqnarray}
\!\!\!\!\!\!\!\!\!\!\!\!\!\! w\i\t  &\!\!\! \triangleq \!\!\! & \text{Pr}\big(m\i\t|\Y\t\big)  \nonumber \\
\!\!\!\!\!\!\!\!\!\!\!\!\!\! &\!\!\!=\!\!\!& \frac{\text{Pr} \big(R\t(\s,a)| \Y\pt, m\i \t\big) \text{Pr} \big(m\i \k|\Y\pt\big) }
{\sum_{j=1}^{M_{\text{KF}}} \text{Pr} \big(R\t(\s,a)| \Y\pt, m\t^{(j)}\big) \text{Pr}\big(m\t^{(j)}|\Y\pt \big)}, \label{eq:modelProb3}
\end{eqnarray}
where $\Y\pt$ shows the reward sequence $\{R_1(\s,a),R_2(\s,a), \ldots, R\pt(\s,a) \}$ and the denominator is just a normalizing factor to make sure that $\text{Pr}(m\i\t|\Y\t)$ is an appropriate probability density function (PDF). Term  $\aL\i\t \triangleq \text{Pr} \big(R\t(\s,a)|\Y\pt, m\i \big)$ in the nominator is the likelihood function of the filter $i$, which is calculated as a PDF of measurement residual of KF  ($\epsilon\t =R\t(\s,a)-\h\t (\bt^a_{t|t-1})\i $) as follows
\begin{eqnarray}
 \aL\i\t &=& \text{Pr}\left(R\t(\s,a)|(\bt^a_{t|t-1})\i, (P^{N}\t)\i \right)  \\
&=& \frac{1}{\sqrt{\text{det}\big[2\pi Z\t\i\big]}}.e^{\frac{-1}{2}\epsilon\t^T\big(Z\t\i\big)^{-1}\epsilon\t},   \nonumber
\end{eqnarray}
where $ Z\t\i = \h\t (\bm{\Pi}^a_{t|t-1})\i \h\t^T +(P^{N}\t)\i$.
Eq.~\eqref{eq:modelProb3} is, therefore, reduced to
\begin{eqnarray}\label{eq:core2}
w\i\t &\!\!=\!\!& \frac{w\i\pt \aL\i\t}{\sum_{j=1}^{M_{\text{KF}}} w^{(j)}\pt \aL^{(j)}\t}.
\end{eqnarray}
The initial value of the weights are set to $w\i_0=1/M_{\text{KF}}$ for $i=1, 2, \dots, M_{\text{KF}} $. The posteriori estimate ${\bt}^a\t$ and its error covariance $\bm{\Pi}^a\t$ are then obtained as
\begin{eqnarray}
\!\!\!\!\!\!\!\!\!\!\!\!\!\!\!\bt^a\t &\!\!\!\!=\!\!\!\!& \sum_{i=1}^{M_{\text{KF}}}w\i\t (\bt^a\t)\i, \\
\!\!\!\!\!\!\!\!\!\!\!\!\!\!\! \bm{\Pi}^a\t &\!\!\!\!=\!\!\!\!& \sum_{i=1}^{M_{\text{KF}}} w\i\t \bigg( \left(\bm{\Pi}_{t}^a \right)\i + \left( ({\bt}^a\t)\i-\bt^a\t \right) \left(({\bt}^a\t)\i-\bt^a\t \right)^T \bigg).
\end{eqnarray}
%
\section{Experimental design}
\label{sec:design}
This section provides additional information about our experiments in the main manuscript.
\subsection{Initialization of $\SR$ parameters }  \label{sec:parameters}
As previously discussed,  $\SR$ requires first training on the source task in the transfer learning setting. The parameters of $\SR$ should be therefore initialized for learning the source task. This section details the parameters selection/initialization for training $\SR$ on the source tasks.

Finding a proper shape parameter for radial basis functions (RBFs) is generally difficult. The centers of RBFs are typically distributed evenly along each dimension of the state vector, leading to $L=(O_{\text{RBF}})^D$ centers for $D$ state variables and order $O_{\text{RBF}}$ given by the user. The variance of each state variable $(\sigma^2_{\text{RBF}})$ is often initialized to $\frac{2}{O_{\text{RBF}}-1}$. It has been shown that RBFs only generalize local changes in one area of the state space and do not affect the entire state space~\cite{pesce2020radial}. Therefore, $\Sig^{(i)}_0$ is a $D \times D$ diagonal positive definite matrix with the entries of $\sigma^2_{\text{RBF}}=\frac{2}{O_{\text{RBF}}-1}$.
 The learning rates for updating mean and covariance of RBFs-based features vector $\bm{\phi}(\s)$  through minimization of loss function $J\t$ are selected by trial and error. The discount factor denoted by $\gamma$ affects how much weight is given to the future rewards in the value function. Commonly (as is the case in the implemented environments), a large portion of the reward is earned upon reaching the goal. To prioritize this final success, we expect an acceptable $\gamma$ to be close to 1.

To use the KFs implemented for learning $\bt^a$ and $\F^a$, the parameters of two KFs have to be initialized: the covariances of the observation noises, the priors and the covariances of the process noises. The priors $\left<\F^a_0 ,\bt_0^a\right>_{a \in \mA}$ should be initialized to the values close to the ones that look optimal or to a default value (i.e., the zero matrices and vectors, respectively). The prior covariances $\left< \S_0^a, \bm{\Pi}_0^a \right>_{a \in \mA}$,  respectively, show the uncertainty in the prior guess of $\left<\F^a_0 ,\bt_0^a\right>_{a \in \mA}$, the lower the more certain. The process noise covariance of a KF is a design parameter. If some knowledge about non-stationarity is available, it can be used to choose this matrix. However, such knowledge is generally difficult to obtain in advance. In this work, the process noise covariances $\P^{\bm{\omega}}$ and $\Sig^{\A}$ are considered time-invariant and chosen by trial and error. They may not be the best ones, but orders of magnitude are correct. Finally, the measurement noise covariance of a KF is one of the most important parameters to be identified. We select this parameter for the KF used for learning $\bt^a$  (i.e., $P^N$) from the following potential range:
{ \begin{eqnarray}
P^{N} \in \{0.01,0.1, 0.2, 0.5, 1, 2, 5, 10, 20, 50\}.
\end{eqnarray}}\normalsize
 The measurement noise covariance $\Sig^{\bm{B}}$ is also selected by trial and error.
\subsubsection{Continuous navigation task}
Since the number of state variables for continuous navigation tasks ($\s=[x,y]^2$) is $D=2$ and we select $O_{\text{RBF}}=4$, $L=16$. The parameters of RBFs for $j=\{1,2,..., 16 \}$ are initialized as
{ \begin{eqnarray}
\!\!\!\!\!\!\!\!\! \um_0^{(j)} &\in& \{ 0.2, 0.4, 0.6, 0.8\}\times \{ 0.2, 0.4, 0.6, 0.8\}, \\
\!\!\!\!\!\!\!\!\! \Sig_0^{(j)} &=& \frac{2}{3} \I_2 \quad \text{for} \, i=\{1,2,..., 16 \},
\end{eqnarray}} \normalsize
where $\I_2$ denotes an identity matrix of dimension of $2\times 2$.

{
\begin{table*}[!t]
\caption{\small Hyper-parameters of the proposed $\SR$ framework for learning task A (continuous navigation task) and task 1 (combination lock task).}\label{Table:param_nav}
\centering
\begin{tabular}{l l|l l}
\hline
 {Hyperparameters\!\!\!} & {\!\!\! Symbol} & {Continuous navigation task} & {Combination lock task}\\
\hline
Discount factor & $\gamma$ & $0.95$  & $0.99$\\
Dimension of the feature vector $\bm{\phi}$ & $L$ & $16$ & $25$\\
Initial estimate of $\bt^a$ for action $a$ & $\bt_0^a$ & $0$ & $0$ \\
 Number of KFs in the multiple-model adaptive structure & $M_{\text{KF}}$ & $1$ & $1$\\
Process noise covariance for learning $\bt^a$ & $\P^{\bm{\omega}}$ & $0.01 \I_{16}$ & $0.01 \I_{25}$\\
Measurement noise variance for learning $\bt^a$  & $P^{N}$ & $ 0.2$ & $ 0.5$\\
Initial posteriori  covariance for learning $\bt^a$  & $\bm{\Pi}^a_0$ & $0.1 \I_{16}$ & $\I_{25}$\\
Initial estimate of $\F^a$ for action $a$ & $\F^a_0$ & $0.02 \I_{16}$ & $0.5 \I_{25}$\\
 Process noise covariance for learning $\F^a$  & $\Sig^{\A}$ & $0.6 \I_{256}$ & $0.5 \I_{625}$\\
Measurement noise covariance for learning $\F^a$  & $\Sig^{\B}$ & $ \I_{16}$ & $ \I_{25}$\\
Initial posteriori covariance for learning $\F^a$   & $\S_0^a$ & $5 \I_{256}$ & $3 \I_{625}$\\
Initial mean of $j$th RBF  & $\bm{\mu}_0^{(i)}$ & $\in \{0.2, 0.4, 0.6, 0.8\}^2$  & $\in \{ 0, 1.2, 2.4, 3.6, 4.8 \}^2$\\
Learning rate for $\bm{\mu}^{(j)}$ & $\alpha_{\bm{\mu}}$ & $0.001$ & $0.01$\\
Initial covariance of $j$th RBF  & $\Sig_0^{(i)}$ &  $\frac{2}{3}\I_2$ &  $0.5\I_2$ \\
Learning rate for $\Sig^{(j)}$ & $\alpha_{\Sig}$ & $0.001$  & $0.005$\\
Epsilon in MB-SR($\epsilon$) algorithm & $\epsilon$ & $0.2$  & $0.02$\\
\end{tabular}
\end{table*}
}
\subsubsection{Combination lock task }
For the combination lock (source task 1) designed in the manuscript, since the digit from the right dial is irrelevant for the reward or transition dynamics prediction, it is ignored; hence, the first and second dimensions of state $\s=[x_1,x_2,x_3]^T$ are considered for constructing the feature vectors, i.e., all states $\s$ with the same left and middle dials digits are mapped to the same feature vector $\bm{\phi}(\s)$. We select $O_{\text{RBF}}=5$ for $x_1$ and $x_2$. Therefore, $L=25$, and for  $j=\{1,2,..., 25 \}$:
{ \begin{eqnarray}
\!\!\!\!\!\!\!\!\!\!\!\!\! \um_0^{(j)} & \!\!\!  \in \!\!\! & \{ 0, 1.2, 2.4, 3.6, 4.8\}\times \{ 0, 1.2, 2.4, 3.6, 4.8 \}, \\
\!\!\!\!\!\!\!\!\!\!\!\!\! \Sig_0^{(j)} &\!\!\!  = \!\!\! & 0.5 \I_2.
\end{eqnarray}} \normalsize
%
%

 We list all the hyper-parameters in this experiment in Table~\ref{Table:param_nav}.
%
\section{Additional experimental results: Transfer with state feature changes} \label{sec:appendix_additional}
In this section, we present the last simulation result illustrating that the learned models in $\SR$ can only generalize to changes that approximately preserve the states' equivalences (feature vectors) but differ in their transitions and rewards.

\textbf{Test task:} We consider test task 3 shown in Fig.~\ref{fig:lock3} where the middle dial rotates randomly, and the right dial becomes relevant
for maximizing reward. In test task 3, setting the left dial to two and the right dial to three is
rewarding, and simulations are started by selecting the left dial to two and the right dial to four. As task 3 is achieved from task 1 by changing its rewarding combination and the rotation direction of the left dial, task 3 differs from the source task 1 in both the dynamics and reward function.

\textbf{Experimental setup:}
To assess if the information of our framework, $\SR$, about the source task 1 can be reused to accelerate learning the test task 3, the learned parameters $\{\bm{\phi}(\s), \bm{S}^a\}_{\s \in \mS}, \{\F^a, \}_{a \in \mA}$, $\{\bt^a, \bm{\Pi}^a \}_{a \in \mA}$, $\F^{\pi}$, and $\bt^{\pi}$ from task 1 are transferred to task 3 for initialization of $\SR_{\text{test}}$.

\textbf{Results:} Fig.~\ref{fig:task3_steps} illustrates the episode length for learning the test task 3. The optimal policy must learn task 3 in $4$ time steps. As we expected, the transfer of knowledge in TD-based algorithms such as SU~\cite{SU} and MB Xi~\cite{reinke2021xi} cannot improve the sample efficiency of task 3, which has different dynamics than task 1. However, it is inferred from Fig.~\ref{fig:task3_steps} that transfer of knowledge to task 3 in both $\SR$ and AdaRL~\cite{huang2021adarl} leads to negative transfer because the right dial is important for predicting expected reward sequences in test task 3 while task 1 ignores the right dial. Therefore, the weights and feature vectors learned in source task 1 are no longer reward and transition predictive in test task 3 and cannot be reused without modification. This is in contrast to test task 2, where the right dial still spins at random and is irrelevant for the reward and transition prediction. \\
Therefore, feature vectors learned by enforcing the known relationships between feature vectors $\bm{\phi}\t(\s)$, $R\t(\s,a)$, and $\mathbb{E}_{P}[\bm{\phi}(\s\nt)|\s,a]$ in Eqs.~\eqref{Eq:LAM2} and~\eqref{Eq:LAM1} encode which state information is relevant for predicting reward sequences and transitions. Because they only model this aspect of an MDP, transfer algorithms such as AdaRL and our proposed framework that utilize the MDP model afford generalization across tasks that preserve these state equivalences but have different transitions and rewards.
\begin{figure}[btp]
\centering
\includegraphics[scale=0.55]{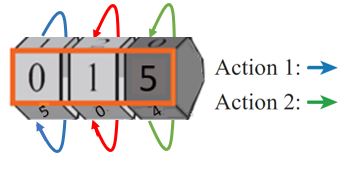}
\caption{Test task 3, where the middle dial rotates at random, and the right dial is important for finding the optimal policy. Therefore, the feature vectors learned from the source task 1 do not apply to task 3.} \label{fig:lock3}
\end{figure}
\begin{figure}[tp]
\centering
\includegraphics[scale=0.43]{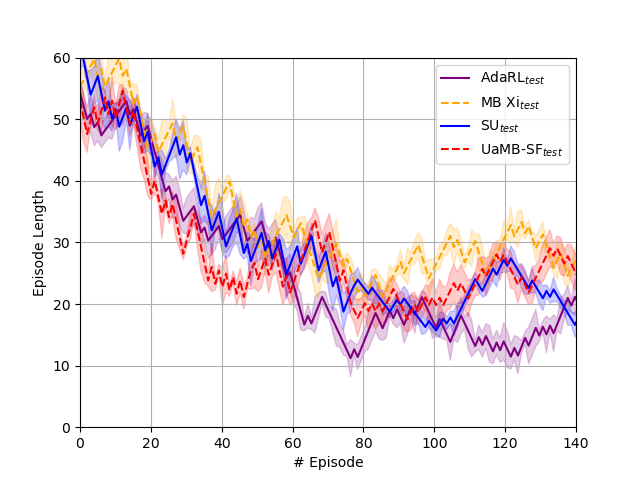}
\caption{Average episode lengths plus standard deviations of task 3, averaged over 20 runs. The state feature vectors learned in source task 1 are no longer reward and transition dynamics predictable in the test task 3.} \label{fig:task3_steps}
\end{figure}
\subsection{Additional details for reproducibility}
\begin{table}[bp]
\caption{Software and hardware configuration used to run all experiments}\label{Table:3}
\centering
\begin{tabular}{l |l }
\hline
\textbf{\!\!\!\ Component\!\!\!} & \textbf{Description} \\
\hline
 Operating system & Windows 10 Pro  \\
 Python & $3.8.5$ (Anaconda) \\
Tensorflow & $2.3.1$ \\
 System Memory & $32$ GB \\
 Hard Disk & $876$ GB \\
 CPU & Intel Xeon(R) W-2265 @ $3.50$GHz \\
 GPU & Nvidia RTX $2080$ \\
\end{tabular}
\end{table}
The experiments were run on a Dell Precision 5820 Tower, whose specifications are provided in Table~\ref{Table:3}. Please note that TensorFlow has only been used to implement the SU algorithm~\cite{SU}. The majority of the computation time in our proposed algorithm is allocated to the matrix-based KF formulation for the estimation of weight matrix $\F^a$, which is computed in $O( L^3 )$ per iteration.

\newpage
\end{appendices}

\bibliography{Hybrid_Model-based_Successor_Representation}
\end{document}